\documentclass[acmlarge]{acmart}
\AtBeginDocument{%
  \providecommand\BibTeX{{%
    \normalfont B\kern-0.5em{\scshape i\kern-0.25em b}\kern-0.8em\TeX}}}

\setcopyright{acmcopyright}
\copyrightyear{2018}
\acmYear{2018}
\acmDOI{10.1145/1122445.1122456}

\acmConference[Woodstock '18]{Woodstock '18: ACM Symposium on Neural
  Gaze Detection}{June 03--05, 2018}{Woodstock, NY}
\acmBooktitle{Woodstock '18: ACM Symposium on Neural Gaze Detection,
  June 03--05, 2018, Woodstock, NY}
\acmPrice{15.00}
\acmISBN{978-1-4503-XXXX-X/18/06}

\usepackage{longtable}
\usepackage{multirow}
\usepackage{graphicx}
\usepackage{threeparttable, tablefootnote}
\usepackage{svg}
\usepackage{caption}
\usepackage{subcaption}

\begin{document}
\title{A Survey on Societal Event Forecasting with Deep Learning}

\author{Songgaojun Deng}
\email{sdeng4@stevens.edu}
\author{Yue Ning}
\email{yue.ning@stevens.edu}
\affiliation{
  \institution{Stevens Institute of Technology}
  \streetaddress{1 Castle Point Terrace}
  \city{Hoboken}
  \state{NJ}
  \country{USA}
  \postcode{07040}
}


\begin{abstract} 
Population-level societal events, such as civil unrest and crime, often have a significant impact on our daily life. Forecasting such events is of great importance for decision-making and resource allocation. 
Event prediction has traditionally been challenging due to the lack of knowledge regarding the true causes and underlying mechanisms of event occurrence.
In recent years, research on event forecasting has made significant progress due to two main reasons: (1) the development of machine learning and deep learning algorithms and (2) the accessibility of public data such as social media, news sources, blogs, economic indicators, and other meta-data sources.
The explosive growth of data and the remarkable advancement in software/hardware technologies have led to applications of deep learning techniques in societal event studies. 
This paper is dedicated to providing a systematic and comprehensive overview of deep learning technologies for societal event predictions. We focus on two domains of societal events: \textit{civil unrest} and \textit{crime}. We first introduce how event forecasting problems are formulated as a machine learning prediction task. Then, we summarize data resources, traditional methods, and recent development of deep learning models for these problems.
Finally, we discuss the challenges in societal event forecasting and put forward some promising directions for future research.
\end{abstract}




\maketitle

\section{Introduction}
Societal events can be broadly categorized into offline events (e.g., strikes, protests, and robberies) and online events (e.g., activism, petitions, and hoaxes on online platforms~\cite{zhao2021event}). 
This paper focuses on studying offline societal events that occur in specific places and times and affect society in different ways.
Understanding such events and their recurring patterns is an urgent issue for many stakeholders such as investors and suppliers. 
Different from retrospective studies such as event detection~\cite{yang1998study} and summarization~\cite{chakrabarti2011event}, event forecasting focuses on anticipating events in the future based on historical event patterns.
Accurate prediction of future events is conducive to effective allocation of public resources, reducing economic loss and social damage. 
Generally, it can bring enormous benefits to society and individuals, such as effective disaster response and socio-economic growth.
In this survey, we focus on summarizing the literature on societal event prediction, where events are initiated by humans and affect society on a large scale. 
Specifically, we organize related work on the prediction of civil unrest and criminal activities that have widespread social impacts from individuals to cities and nations~\cite{braha2012global}.
 
With the availability of mass media sources, data-driven approaches have been widely studied in various fields, including civil unrest~\cite{deng2019learning} and criminal activities~\cite{huang2018deepcrime}.
In these studies, researchers attempt to develop predictive analytic techniques based on historical data of events and other relevant sources to predict future events.
The methods studied are from a variety of fields, such as statistics, data mining, machine learning, and deep learning.
Nowadays, the effective collection of good-quality data, the popularity of high-performance computing equipment~\cite{flouris2017issues}, and the accelerated development of the field of machine learning and deep learning have led to substantial progress in event prediction research.
However, there are still many challenges present in addressing this problem:
\begin{itemize}
    \item \textbf{Leveraging Heterogeneous Data.} As the availability of various open-source data continues to increase, researchers have begun to resort to heterogeneous data to develop predictive models~\cite{mengleveraging,stec2018forecasting,wang2020deep}. 
    Societal events occur in a dynamic social environment, and their corresponding key information can publish in various forms, such as traditional news reports, social media, and government official reports. 
    It brings unique challenges of efficiently processing and learning from heterogeneous data to make accurate event predictions. 
    \item \textbf{Studying Dependencies in Societal Events.} 
    Societal events exhibit geographical properties and also have a high degree of temporal dependency~\cite{zhao2015spatiotemporal}. Modeling event contextual information requires an in-depth investigation of the spatiotemporal dependencies of events.
    Traditional methodologies have shown limitations in modeling complex event data~\cite{olligschlaeger1997artificial,smith2017predicting}, which encourages the development of advanced models to address this challenge.
    \item \textbf{Interpreting Event Predictions.} 
    Event prediction models enable human users to automatically forecast future events of interest without manually inspecting historical data.
    Along with event forecasting, interpreting the prediction results is an equally important task, as it can assist practitioners in understanding prediction results and making reasonable and practical decisions. 
    Data-driven event prediction models with adequate interpretability are more conducive to supporting human users in event analysis and decision making than black box models.
\end{itemize}
 
Over the past few decades, a large amount of research has been devoted to the development and application of event prediction techniques in response to the above challenges.
Deep learning has gained tremendous success in many applications such as computer vision, speech recognition, and natural language processing, given its superiority in improving accuracy when trained with sufficient data. 
Recently, there has been a surge of deep learning approaches in the societal event domain.
The purpose of this survey is to summarize deep learning technologies for event forecasting and to highlight the open problems and new possibilities for societal event prediction.

\subsection{Previous Work and Contributions}
There are several other surveys on societal event predictions.
\citeauthor{phillips2017using}~\cite{phillips2017using} provided a literature review that examines the problems and techniques for predictive analysis using social media data. Recently, \citeauthor{zhao2021event}~\cite{zhao2021event} provided a systematic survey of existing data-driven event prediction methods, covering challenges, techniques, applications, evaluations, and open problems.
As event prediction methods are typically motivated by specific application areas, some surveys instead focus on event technologies in particular domains.
\citeauthor{ganar2016prediction}~\cite{ganar2016prediction}
summarized studies of civil unrest prediction in social media analysis based on keyword filtering.
For predictive studies of crime,
\citeauthor{mookiah2015survey}~\cite{mookiah2015survey} discussed crime research that considers various crime-related variables and examined the impact of these variables on crime prediction.
\citeauthor{kounadi2020systematic}~\cite{kounadi2020systematic} provided an overview and evaluation of the state-of-the-art technologies in spatial crime prediction, with a focus on research design and techniques. 
\citeauthor{butt2020spatio}~\cite{butt2020spatio} investigated detection and prediction methods for crime hotspots, mainly covering papers published in the recent literature.

In contrast to previous work, this survey is structured around problems, data, and deep learning models for civil unrest and crime forecasting. 
Civil unrest and crime are influential societal events and share some common characteristics. 
Both civil unrest and crime are human-initiated events, largely depending on evolving social environments and affecting humans and their life to a large extent. 
We seek to provide a systematic survey that bridges the prediction of such influential societal events with the development of deep learning and point out possible directions for research and development.
The main contributions of this work can be summarized as follows:
\begin{itemize}
    \item   \textbf{A Comprehensive Summary of Data Resources} 
   Developing data-driven approaches for predicting societal events relies heavily on big data. 
In this work, we organize and categorize the event datasets that have been investigated in existing work.
 As data are the basis for data-driven modeling of societal events, this summary will help researchers and human experts in several ways, such as selecting critical features for societal event analysis.
    \item \textbf{A Systematic Review of Existing Deep Learning Methods.} 
    This survey categorizes existing deep learning
    methods for predicting two types of societal events (i.e., civil unrest and crime). 
    We analyze the characteristics of deep learning-based technologies for each event domain.
    Traditional statistical and machine learning methods are also summarized, with their limitations discussed.
    The methodological review aims to help researchers systematically understand the advantages and achievements of various deep learning techniques in societal event prediction.
    \item  \textbf{A Discussion of Ongoing Challenges and Future Prospects.} 
    This paper provides an overall outline of the current research status by discussing ongoing research challenges and future directions. With this discussion, we aim to provide comprehensive insights to practitioners in related fields and advance the studies of societal event prediction.
\end{itemize}

 \subsection{Outline}
The rest of this paper is organized as follows.
We first present the usage of mathematical notations in this survey in Section~\ref{notation}.
Section~\ref{civil} presents the problem formulation of civil unrest prediction, along with the introduction of data resources, traditional methods, and a comprehensive review of deep learning models developed for civil unrest prediction. 
Section~\ref{crime} discusses crime prediction, followed by the same outline as in Section ~\ref{civil}.
In Section ~\ref{tech}, we summarize the deep learning technologies studied in civil unrest and crime prediction.
Section ~\ref{challenge} lists open problems and suggestions for future research directions. This survey concludes with a summary in Section ~\ref{conclusion}.

\subsection{Notation}\label{notation}
To ensure the readability of the survey, we use consistent symbols when describing variables. Calligraphy capital letters represent general variables. For example, $\mathcal{X}$ denotes input data which could be in any format, and $\mathcal{Y}$ is the generic target variable, which can be a binary or continuous value. Bold lowercase letters indicate vectors (e.g., $\mathbf{x}$), bold uppercase letters indicate matrices (e.g., $\mathbf{X}$), and heavier uppercase letters signify tensors/higher dimensional collections of data (e.g., $\pmb{\mathbf{X}}$). 
We use the letters $W,b$ to represent learnable model parameters.
For simplicity and consistency, we will use the letter $f$ (e.g., $f(\cdot)$) in some cases to represent the components in a deep learning model such as a feedforward neural network.
$\sigma$ represents the sigmoid function, $\odot$ represents the Hadamard product, and $[;]$ signifies concatenation.

\section{Civil Unrest Prediction}\label{civil}
Civil unrest, also known as civil disorder, is an activity arising from a mass act of gathering. 
In such activities, participants express their protest against things that affect their lives and for which they assume that governments or other entities have a responsibility (e.g., cost of urban transportation, poor infrastructure, etc.).
Civil unrest events include protests, strikes, and demonstrations for different reasons.
Anticipating such events can help stakeholders (e.g., investors and suppliers) understand the dynamic patterns of these events and improve resource allocation.
This section starts with the formal problem formulation of civil unrest prediction followed by an introduction of available data resources in this research area.
In the following, we summarize traditional civil unrest forecasting methods and provide a more detailed description of advanced deep learning approaches categorized by the characteristics of model structures.

\subsection{Problem Formulation}
Event forecasting can be expressed as a supervised learning task in machine learning, which aims at learning a function that maps an input to an output based on input-output pairs.  
It infers a function from labeled training data consisting of a set of training instances, where the input is historical data collected before the timestamp of the output variable.
Before going into the problem definition of societal event forecasting,  we first introduce the preliminaries, including terminologies and mathematical notations.

\textbf{Historical Data}. Suppose there are $L$ locations (e.g., cities, states) of interest, and each location $l$ can be represented by a set of features used for prediction. We divide features into two categories, static and dynamic.\footnote{Dynamic features are usually involved in civil unrest prediction studies, while static features are sometimes not considered.
} Static features such as population and political ideology remain constant or change slowly over a long time, while dynamic features such as frequency of events or number of tweets expressing ``angry'' emotions are updated for each time interval $t$ (e.g., day, week).
Let $\mathcal{S}_l$ denote the set of
static features of location $l$, and $\mathcal{X}_{t,l}$ be the collection of dynamic features for location $l$ at time $t$.
The collection of dynamic features from location $l$ within an historical window (i.e., observing time window) with size $k$ up to time $t$ can be represented as $\mathcal{X}_{t-k+1:t,l} = (\mathcal{X}_{t-k+1,l},...,\mathcal{X}_{t,l} )$.

\textbf{Ground-truth Event Occurrence}. A target variable $\mathcal{Y}_{t^*,l}$ indicates the occurrence of a future civil unrest (e.g., protest) for each location $l$ at time $t^*$. Note that $t^*$ can be either a time point ($t+\Delta$) or a time window in the future ($t+\Delta:t+\Delta+\delta $). 
$\Delta \geq 1$ is the \textbf{lead time} that denotes the number of time steps in advance for a prediction. We use $\delta \geq 0$ to denote the \textbf{lead time window} that represents whether a civil unrest event will occur between time $t+\Delta$ and $t+\Delta+\delta $.

\begin{definition}
\textbf{Civil Unrest Prediction as Classification}. 
Learn a classifier $f(\mathcal{S}_l,\mathcal{X}_{t-k+1:t,l})\rightarrow \mathcal{Y}_{t^*,l}$ that maps an input including static and dynamic features, to a categorical civil unrest variable $\mathcal{Y}_{t^*,l} \in \{0,1,..., M-1\}$ at the future time $t^*$ for the target location $l$.
If there is only one category of events to be predicted, $\mathcal{Y}_{t^*,l}=1(0)$ indicates that events of interest will (not) occur at location $l$ at time $t^*$. If there are multiple categories of events to be predicted at the same time ($M>2$), 
this problem becomes a multi-label classification problem where the target value indicates the occurrence of multiple classes (e.g., multiple subtypes of events), i.e., $\mathcal{Y}_{t^*,l} = \{0,1\}^{M}$. 
\end{definition}

\begin{definition}
\textbf{Civil Unrest Prediction as Regression}.
Given static features $\mathcal{S}_l$ and dynamic features $\mathcal{X}_{t-k+1:t,l}$,  the goal is to learn a regressor $f(\mathcal{S}_l,\mathcal{X}_{t-k+1:t,l})\rightarrow \mathcal{Y}_{t^*,l}$ that maps the input to a numerical civil unrest indicator $\mathcal{Y}_{t^*,l} \in \mathbb{R}$ at the future time $t^*$ for the target location $l$. In general,
$\mathcal{Y}_{t^*,l}$ refers to normalized number of event frequencies.  
\end{definition}

\subsection{Data Resources}
In this section, we summarize the event data sources utilized in civil unrest prediction studies. Event data are typically used as the ground-truth of event occurrence and historical input data. We also discuss some open-source indicators that have been used as historical input data for civil unrest prediction.
\begin{table}[]
 \caption{ A summary of event datasets used for civil unrest prediction. All listed datasets are labeled with geolocation information. The start/end time in Temporal Coverage indicates the earliest/latest time of data collection. ``/'' means unavailable. 
 We use a horizontal line to separate the human-encoded data (top) from the machine-encoded data (bottom).
}
 \label{tab:civil-data}
\small
\scalebox{0.9}{\begin{tabular}{p{1.6cm} p{2.6cm} p{1.2cm} p{1.8cm} p{5.1cm} p{1.2cm} p{1.0cm}}
\toprule
\textbf{Dataset}  & \textbf{Mendate}                                                                                                                                   & \textbf{Temporal coverage}   & \textbf{Spatial coverage}                & \textbf{Event definition}                                                                                                                                                                                               & \textbf{Coding process} & \textbf{Open sourced} \\
\midrule
COPDAB~\cite{azar1980conflict}   & Political events                                                                      & 1948-1978                                                     & Near global                                                  & /                                                                                                                                                                                                   & Human          & Yes          \\
WEIS~\cite{mcclelland1978world}     & Political events                                     & 1966-1978                                                     & Global                                                         & /                                                                                                                                                                                                   & Human          & Yes          \\
SPEED~\cite{nardulli2011event}    & Social, political and economic events                                                  & 1945-2008                                                     & Global                                                         & /                                                                                                                                                                                                   & Human          & Yes          \\
SCAD~\cite{salehyan2012social}     & Social conflict events & 1990-2017                                                     & Africa and Latin America          & /                                                                                                                                                                                                   & Human          & Yes          \\
ACLED~\cite{raleigh2015armed}    & Political
violence and demonstrations                                                                           & 1997- & Near global                                                    & A single altercation where force is often used by one or more groups to a political end                                                                              & Human          & Yes          \\
\citeauthor{daly2012organizational}~\cite{daly2012organizational}  & Violent events/Rebellion & 1964-1984 & Colombia & The entire repertoire of violence and threats of violence that suggest the presence of a militarized organization  &  Human & No\\
\citeauthor{urdal2012explaining}~\cite{urdal2012explaining} & Civil unrest events  & 1960-2009 & Asia and Sub-Saharan Africa & / & Human & No\\

KOSVED~\cite{schneider2013accounting}   & One-sided violence                                                                                          & 1991-2008                                                     & Africa and Europe   & Violent acts perpetrated by an organized group, directed against a group of unarmed non-combatants during, shortly before, or after a conflict       & Human          & Yes          \\
GTD~\cite{lafree2007introducing}      & Terrorism                                                                                                                    & 1970-2019                                                     & Global                 & The threatened or actual use of illegal force and violence by a non-state actor to attain a political, economic, religious, or social goal through fear, coercion, or intimidation                  & Human          & Yes          \\
UCDP GED~\cite{sundberg2013introducing} & Organized violence                                                                            & 1989-2020                                                     & Near global & An incident where armed force was used by an organised actor against another organized actor, or against civilians, resulting in at least 1 direct death          & Human &  Yes          \\
GSR~\cite{ramakrishnan2014beating}      & Civil unrest events                                                                                                            & /                                                             & Latin America                                                  & Organized by an independent evaluation team (MITRE) by surveying newspapers for reportings of civil unrest                                                                                          & Human          & No    \\      
\midrule
KEDS~\cite{schrodt1994political}     & Political events                                                                                                         & 1979-1997                                                     & Middle East, Balkans, and West Africa                          & Use dictionaries of nouns and verb phrases to code the actors and events                                                                                                                     & Machine        & Yes          \\
Phoenix~\cite{illinoisdatabankIDB-2796521}  & Political events                                                  & 1945-2019                                                     & Global                                                         & CAMEO methodology                                                                                                                                                                                              & Machine        & Yes          \\
GDELT~\cite{leetaru2013gdelt}    & Various physical activities, including political and non-political events                                                                               & 1979-                                                         & Global                                                         & CAMEO  methodology                                                                                                                                                                                             & Machine        & Yes          \\
ICEWS~\cite{DVN28075_2015}    & Political events                                                               & 1995-                                                         & Global                                                         & CAMEO methodology                                                                                                                                                                                              & Machine        & Yes          \\
AutoGSR~\cite{saraf2016embers}  & Civil unrest events                                                                                                           & /                                                             & Latin Americ                                                   & Use the developed automatic event coding system to classify articles as protests or non-protests                                                                                                    & Machine        & No           \\
\bottomrule
\end{tabular}}
\end{table}

\subsubsection{Civil Unrest Event Related Data} 
Since the last century, various event data projects with different data collection, coding, and analysis processes have emerged.
We organize event data related to civil unrest in Table~\ref{tab:civil-data}. 
According to the coding process of event data, we divide these data into human-coded and machine-coded data.
Human-coded data depend on human research teams with specific knowledge of the local context, while machine-coded data rely entirely on automated event encoding systems.

\textbf{Human-Encoded Events.}
Many human-coded event datasets have been developed and maintained, which allow researchers to build forecasters at specific sub-state geographic units~\cite{weidmann2010predicting,schneider2012dynamics,mengleveraging}.
In the earliest days, due to technological limitations (i.e., the lack of electronic articles and computational power), the World Event Interaction Survey (WEIS)~\cite{mcclelland1978world} and the Conflict and Peace Data Bank (COPDAB)~\cite{azar1980conflict} projects hire human analysts to physically collect newspaper clippings, press reports, and summary accounts from Western news sources to obtain news stories. These projects focus on daily international and domestic events or interactions.
Analysts coded events in a multi-column format based on their subjective judgment. 
The event record mainly includes the date of the event, actors involved in the event, sources from which the information about the event was gathered, issue area(s), and textual description. 

Later, there were more event data projects focused on specific areas.
The Social, Political, Economic Event Database (SPEED) Project~\cite{nardulli2011event} is a technology-intensive effort to extract event data from a global archive of news reports covering the Post WWII era. This project develops more than one hundred attributes to encode each event, excluding text descriptions.
\citeauthor{salehyan2012social} developed an event dataset, Social Conflict in Africa Database (SCAD)~\cite{salehyan2012social}, which contains instances of protests, riots, strikes, government repression, communal violence, and other forms of unrest that happened mainly in Africa. 
The Gold Standard Report (GSR) is a collection of human classified civil unrest news reports from the most influential newspaper outlets in Latin America~\cite{ramakrishnan2014beating}.
The Armed Conflict Location and Event Data Project (ACLED)~\cite{raleigh2015armed} is a disaggregated data collection, analysis, and crisis mapping project.
ACLED collects the dates, actors, locations, fatalities, and types of all reported political events (e.g., violence, protests) around the world.
\citeauthor{urdal2012explaining}~\cite{urdal2012explaining} introduced an event dataset on urban unrests at the city level that covers 55 major cities in Asia and Sub-Saharan African countries. 

In addition to political events, some datasets focus on violent events.
\citeauthor{daly2012organizational}~\cite{daly2012organizational} collected violent events at the municipality-month level in Colombia. 
The Global Terrorism Database (GTD)~\cite{lafree2007introducing} provides information on domestic and international terrorist attacks around the world.
The Konstanz One-Sided Event Dataset (KOSVED)~\cite{schneider2013accounting} provides detailed information on the magnitude and locations of one-sided violent events in 20 civil wars. 
The Uppsala Conflict Data Program Georeferenced Event Dataset (UCDP GED)~\cite{sundberg2013introducing} is an event dataset that disaggregates three types of organized violence (state-based conflict, non-state conflict, and one-sided violence) both spatially and temporally.

\textbf{Machine-Encoded Events.} 
Manual approaches began to be replaced with automated coding with the first iteration of the Kansas Event Data Set (KEDS)~\cite{schrodt1994political} project in the late 1980s. KEDS uses the automated coding of English-language news reports to generate political event data. These data are used in statistical early warning models to predict political changes.
The Integrated Crisis Early Warning System (ICEWS)~\cite{DVN28075_2015} includes a database of political events with global coverage.
Similar to ICEWS, the Global Dataset of Events, Location, and Tone (GDELT)~\cite{leetaru2013gdelt} has been developed and compiled a comprehensive list of electronic news sources.
Both GDELT and ICEWS are active automatic systems that identify and classify events from public data following the Conflict and Mediation Event Observations (CAMEO) ~\cite{gerner2002conflict} which is a framework for coding event data.
These two datasets have been extensively studied in various fields.
Some researchers compared ICEWS with GDELT~\cite{ward2013comparing,arva2013improving,Wang1502} and pointed out limitations on local conflict processes that rely too heavily on machine-coded data~\cite{hammond2014using,halkia2020conflict}.
The Historical Phoenix Event Data~\cite{illinoisdatabankIDB-2796521} includes events extracted from
The New York Times, BBC Monitoring's Summary of World Broadcasts, and the CIA's Foreign Broadcast Information Service. It also uses the CAMEO methodology to encode events. 
Given the large scale and extensive spatial coverage of machine-coded data, many researchers have used machine-coded event data to build forecasts for civil unrest~\cite{yonamine2013predicting,qiao2017predicting,qiao2020learning,zhao2015multi,zhao2016multi,zhao2017feature,ning2018staple,gao2018incomplete,zhao2018distant,deng2019learning}.

\subsubsection{Open Source Indicators}
In civil unrest prediction, researchers typically define ground truth data for learning tasks.
In general, historical civil unrest or related events can be used as basic features for modeling future event patterns of interest.
Recent research has incorporated public media data as features for civil unrest forecasting. This type of data is known as Open Source Indicators (OSI). OSI includes traditional media data such as digital newspapers, blogs, and social media data such as posts from Twitter and Facebook. 
These data provide a wealth of background information that helps one understand the social context and public opinion of civil unrest events.
Economic indicators and other meta-data sources have also been explored in this line of research. 
Studies show that exogenous political and economic variables can serve as the necessary underlying drivers of social unrest besides social media~\cite{wolfsfeld2013social,mengleveraging}.
Social and economic features derived from the World Development Indicators (WDI)~\cite{world2016world} and Worldwide Governance Indicators (WGI)~\cite{kaufmann2011worldwide} have been investigated
~\cite{parrish2018crystal}.
Researchers have also utilized Google Trends (GT) to uncover social dynamics associated with behavior that precedes episodes of civil unrest~\cite{manrique2013context}. Google Trends analyzes the popularity of top search queries in Google Search across various regions and languages.

\subsection{Traditional Methods}

Over the years, researchers have leveraged various predictive
techniques for civil unrest predictions ranging from simplest threshold-based methods~\cite{manrique2013context} to more sophisticated methods such as logistic regression and deep neural network. We outline the development of traditional predictive approaches in this field before introducing deep learning-based methods.

As a pioneering study, \citeauthor{radinsky2013mining}~\cite{radinsky2013mining} mined chains of events from massive news archives and proposed a probabilistic method that predicts the likelihoods of future worldwide events of interest.  
\citeauthor{jin2014modeling}~\cite{jin2014modeling} characterized mass protest propagation using a bispace diffusion model comprising a Twitter mentions network and a latent space. 
\citeauthor{chen2014non}~\cite{chen2014non} explored a nonparametric graph scan algorithm to the problem of civil unrest event detection and forecasting using heterogeneous social media graphs.
Subsequently, people began to apply traditional machine learning models to civil unrest prediction.
\citeauthor{kallus2014predicting}~\cite{kallus2014predicting} studied the power of massive online-accessible public data to predict crowd behavior such as protests utilizing random forests.
Logistic regression as a classification technique has been extensively explored in combining heterogeneous data sources for civil unrest forecasting~\cite{ramakrishnan2014beating,cadena2015forecasting,korkmaz2015combining,zhao2015multi,wu2017forecasting}.
Early Model Based Event Recognition using Surrogates (EMBERS)~\cite{ramakrishnan2014beating,muthiah2016embers} is an automated system developed for generating forecasts about civil unrest from massive and multiple data sources. 
This system consists of individual models that leverage specific data sources, including logistic regression, lexical pattern matching, dynamic query expansion, etc. 
More advanced methodologies based on logistic regression have emerged.
Multi-task learning frameworks were incorporated in forecasting spatial-temporal civil unrest events~\cite{zhao2015multi,zhao2016multi,zhao2017feature,ning2018staple,gao2018incomplete,zhao2018distant}. \citeauthor{ning2016modeling}~\cite{ning2016modeling} introduced a multi-instance learning method for forecasting and modeling precursors for civil unrest.
More recently, \citeauthor{zhao2016hierarchical} ~\cite{zhao2016hierarchical} presented a group-Lasso based hierarchical feature learning model to characterize feature dependence, feature sparsity, and interactions among missing values.
Temporal models such as Hidden Markov Models (HMM) were proposed to leverage temporal burst patterns from large-scale digital history-coded events~\cite{qiao2017predicting,qiao2020learning} or Twitter streams~\cite{zhao2015spatiotemporal} to reveal the underlying developmental mechanism of civil unrest events.
Other work considered civil unrest forecasting as a regression task and investigated time series models such as autoregressive model~\cite{weidmann2010predicting,schneider2012dynamics,yonamine2013predicting}.

Although substantial progress has been made, traditional approaches usually require extensive efforts in feature engineering, such as analyzing news reports to obtain keywords as feature variables.
Moreover, these models often consider only simple features and cannot model complex dependencies in the data, which limits the predictive power of such models. 
Next, we discuss deep learning-based models that are capable of capturing complex data dependencies.

\subsection{Deep Learning Models}\label{sec:civil-deep}

\begin{table}[]
\caption{Deep learning based civil unrest prediction techniques. We use abbreviations ``class.'' and ``reg.'' to indicate classification and regression, respectively. 
We use horizontal lines to partition the models with different structural characteristics, corresponding to section~\ref{sec:civil-deep}.
}
\small
\label{tab:civil-models}
\begin{threeparttable}
\scalebox{0.9}{\begin{tabular}{p{2.0cm} p{1.5cm} p{0.8cm} p{1.2cm} p{1.6cm} p{1.8cm} p{1.4cm} p{1.0cm} p{1.0cm} p{1.2cm}}
\toprule
\multirow{2}{*}{\textbf{Technique}}       & \multicolumn{2}{l}{\textbf{Space}}   & \multirow{2}{*}{\textbf{Time}}    & \textbf{Dataset}    &    & \multicolumn{3}{l}{\textbf{Forecasting}}       & \multirow{2}{*}{\textbf{Evaluation}}     \\
\cmidrule(r){2-3}\cmidrule(r){5-6}\cmidrule(r){7-9}
      &     \textbf{Study area }        & \textbf{Scale}         &      & \textbf{Event data}   & \textbf{External data}      & \textbf{Inference}      & \textbf{Task}    & \textbf{Temporal unit} &      \\
\midrule
LSTM~\cite{smith2017predicting}                                                                                               & Afghanistan                              & Country       & 2001-2012                                                   & GDELT                                                            & /                                                         & \# of material conflict events  & Reg.            & Month         & MAE, RMSE, MAPE                 \\
GRU~\cite{parrish2018crystal}                                                                                                        & 158 countries                            & Country       & Mar 2001-Mar 2014 & GDELT                                                            & WDI, WGI                                                  & Occurrence of disruptive events & Binary class. & Month         & AUC, ROC                        \\
LSTM~\cite{halkia2020conflict}                                                                                                        & Libya, Sudan, Egypt, Maldives, Nicaragua & Country       & 1989-2019                         & ICEWS, GDELT                                                     & /                                                         & \# of material conflict events  & Reg.            & Month         & RMSE                            \\
Cov-LSTM~\cite{mengleveraging}                                                                           & Egypt, Jordan                            & Country, city & May 2015-Jan 2019                                           & GSR, ACLED                                                       & WDI, Trading Economics, Wikipedia, Twitter, Google Trends & \# of civil unrest events       & Class.        & Day           & Mercury score\tnote{a}, ROC              \\
\midrule
CA-LSTM~\cite{wang2018unrest} & USA                            & State         & Jan 2012-Jan 2018                                           & GDELT                                                            & /                                                         & \# of material conflict events  & Reg.            & Day           & MSE, MAE                        \\
ActAttn~\cite{ertugrul2019activism}                                           & USA                            & State         & 2014, 2017                                                  & Charlottesville rally, Ferguson protests I, Ferguson protests II & Twitter                                                   & Occurrence of protests          & Binary class. & Day           & AUC, F1                         \\
\midrule
DynamicGCN~\cite{deng2019learning}                            & Thailand, India, Egypt, Russia           & City          & 2010-2016                                                   & ICEWS                                                            & News reports                                               & Occurrence of protests          & Binary class. & Day           & Precision, Recall, F1    \\
Glean~\cite{deng2020dynamic}                         & Nigeria, India, Afghanistan, Russia           & City          & 2012-2016                                                   & ICEWS                                                            & News reports, Knowledge graphs                                               & Occurrence of multiple political events          & Class. & Day           & F1, F2\tnote{b},  Recall    \\
CMF~\cite{deng2021understanding}                            & Thailand, India, Egypt, Russia           & City          & 2012-2016; 2017-2019                                                   & ICEWS, GDELT                                                            & News reports, Knowledge graphs                                               & Occurrence of protests          & Binary class. & Day           & F1    \\
\bottomrule
\end{tabular}}
\begin{tablenotes}\footnotesize
\item[a] The evaluation metric is defined in the IARPA Mercury Challenge.
\item[b] F2 score is similar to F1 score, but assigns less weight to precision and more weight to recall.
\end{tablenotes}
\end{threeparttable}
\end{table}

A number of deep learning-based approaches have been proposed to predict civil unrest events and have demonstrated their advantages in terms of improved predictive capability or interpretability.
We classify existing work based on their methodology into Recurrent Neural Network (RNN), Attention, and Graph Neural Network (GNN). 
We provide an overview of deep learning-based models that focus on the prediction of civil unrest in Table~\ref{tab:civil-models}.

\subsubsection{Recurrent Neural Network Based Approaches} 
Predicting civil unrest events can be considered as a time series prediction problem. 
Recurrent neural networks (such as RNN, LSTM~\cite{hochreiter1997long} and GRU~\cite{chung2014empirical}) are proposed to capture short- and long-term dependencies in time series data, and they have been proven to be more expressive and powerful than traditional methods such as autoregressive models. 
Researchers have employed recurrent neural networks to model temporal dependencies on time series event data for civil unrest prediction.

Suppose that each sample in the time series data for a location consists of a sequence $\mathbf{X}=(\mathbf{x}_{t-k+1},... ,\mathbf{x}_t)$ with length $k$ where the element $\mathbf{x}_t$ is the feature vector at time $t$, e.g., event frequency. 
The sequence is fed into the RNN architecture for predicting the occurrence of the target event $y_{t^*}$ at the future time $t^*$. 
A vanilla RNN model is defined as belows:
\begin{equation}
    \mathbf{h}_t = \tanh(\mathbf{W} [ \mathbf{x}_t; \mathbf{h}_{t-1}] + \mathbf{b}),   
    \label{equ:rnn}
\end{equation}
where $\{\mathbf{h},\mathbf{x}\}_t$ are the hidden states and input feature vector at time $t$ and $\mathbf{h}_{t-1}$ is the hidden state at time $t-1$. $[;]$ is concatenation. $\{\mathbf{W},\mathbf{b}\}$ are model parameters. 
The time-series event prediction using vanilla RNN can be written in the following form: 
\begin{equation}
     \hat{y}_{t^*} = f(\mathbf{h}_{\tau}), \quad \tau \leq t.
\end{equation}
The prediction $\hat{y}_{t^*}$ can be obtained from a feedforward network that takes the hidden states as input, either the last or all hidden states.
The vanilla RNN model typically fails to store information for a longer period of time and has gradient vanishing and exploding problems.
Existing work usually utilizes variants of RNN (e.g., LSTM and GRU) to model time series data, especially when the input sequence is not short.
\citeauthor{smith2017predicting}~\cite{smith2017predicting} and \citeauthor{halkia2020conflict}~\cite{halkia2020conflict} applied LSTM models to event prediction, focusing on material conflict events such as armed attacks and destruction of property.
Unlike RNN models, LSTM includes cell states that can remove or add information to the cell.
Thanks to this structure, LSTM is more accurate on datasets with very long sequences.
GRU is another recurrent neural network that uses fewer training parameters and therefore uses less memory leading to faster training than LSTM. 
~\citeauthor{parrish2018crystal}~\cite{parrish2018crystal} studied a GRU-based multi-feature driven approach to predict disruptive events. Instead of considering the sequence of event counts as input, this work consists of a feature selection process to select social and economic features derived from open-source Indicators.

More recently, \citeauthor{mengleveraging}~\cite{mengleveraging} proposed a model, Cov-LSTM~\cite{mengleveraging}, that combines convolutional layers and LSTM layers. The convolutional layers aim to extract high-level representations of event time series data.
The transformed input features $\mathbf{x}'_t = \text{Conv1D}(\mathbf{x}_t)$ are fed to the LSTM model to predict civil unrest events.

\subsubsection{Attention Based Approaches} 
Several challenges have emerged in RNN-based approaches in civil unrest prediction, including limited ability to accurately predict unrest events and difficulty for users to understand the model behavior. Attention mechanisms were first introduced into a neural machine translation model based on encoder-decoder RNNs~\cite{bahdanau2014neural}.
Attention mechanisms enable dynamically highlighting relevant features of the input data, mimicking cognitive attention in humans. This method enhances the important parts of input data and fades out the rest.
Formally, the attention mechanism takes several hidden vectors $( \mathbf{h}_{t-k+1},...,\mathbf{h}_t )$, and a context vector $\mathbf{c}$ as input, and outputs an attention weight $\alpha_{\tau}, t-k+1 \leq \tau \leq t$ for each hidden vector $\mathbf{h}_{\tau}$ and an attention vector $\mathbf{a}_t$:
\begin{equation}
   \alpha_{\tau} = \text{softmax} \big (\text{score}(\mathbf{h}_{\tau},\mathbf{c}) \big ),  \quad \mathbf{a}_t = \sum_{t-k+1 \leq \tau\leq t} \alpha_{\tau} \cdot \mathbf{h}_{\tau},
  \label{eq:attn}
\end{equation}
where  $\text{score}(\cdot)$ is a scoring function that can be implemented in various forms (e.g., a feed-forward neural network, dot product, etc.) leading to different attention mechanisms, such as additive~\cite{bahdanau2014neural}, dot product~\cite{luong2015effective} or self-attention mechanisms~\cite{vaswani2017attention}. 
The softmax function normalizes the output of the scoring function to the attention weights.
Removing the context vector $\mathbf{c}$ results in a location-based attention, where the attention weight depends only on the target location.

\citeauthor{wang2018unrest} proposed a context-aware attention-based LSTM framework named CA-LSTM~\cite{wang2018unrest} to study different contributions of data points in the time series and to improve the accuracy in predicting civil unrest events. 
They applied an attention layer on top of the LSTM layer to obtain an attention vector. 
To further include trends in the occurrence of events, they explicitly incorporated target event variables in previous historical steps to the attention vector for civil unrest prediction, i.e., $ \hat{y}_{t^*} = f(\mathbf{a}_t, (y_{t-k'},...,y_{t}))$.

\citeauthor{ertugrul2019activism} introduced a hierarchical attention-based spatiotemporal learning approach, ActAttn~\cite{ertugrul2019activism} for predicting the occurrence of future protests and explaining feature importance.
Specifically, the model takes historical data (i.e., Twitter data) from the target (predicted) location $l \in L $, as well as historical data from all locations $L$, as input.
The model consists of a two-level attention module built on LSTM models, which calculates intra-regional (temporal) and inter-regional (spatial) contributions, respectively.
The proposed model can be briefly expressed as bellows:
\begin{equation}
    \hat{y}_{t^*,l} = f(\mathcal{S}_l, \alpha^{\text{temporal}} \mathbf{h}^{\text{temporal}}_{t,l}+\alpha^{\text{spatial}} \mathbf{a}^{\text{spatial}}_t),
\end{equation}
where $\mathbf{a}^{\text{spatial}}_t= \sum_{l \in L} \alpha_l \cdot \mathbf{h}_{t,l}$ is the spatial attention vector, obtained from applying attention on hidden states for all locations. A spatiotemporal attention is applied on the hidden states of the target location $\mathbf{h}_{t,l}^{\text{temporal}}$ and the spatial attention vector $\mathbf{a}^{\text{spatial}}_t$. All hidden states are obtained from LSTM models.
The network can help interpret what features, from which places, have significant contributions to a prediction of protest.

\subsubsection{Graph Neural Network Based Approaches}
Recently, graph neural networks have achieved significant development in the machine learning community and have been applied to a wide range of fields such as computer vision~\cite{shen2018person}, natural language processing~\cite{yao2019graph}, bioinformatics~\cite{rhee2017hybrid}, etc.
These methods have demonstrated superior performance in modeling latent embedding in graph-structured data. 

Graph neural networks (GNNs) are a class of neural networks for prediction tasks on graph-represented data. 
GNNs learn embeddings/hidden features for each node in a graph by using topological information.
In general, one GNN layer utilizes the node embeddings from last layer and the graph structure of a graph:
\begin{equation}
    \mathbf{H}^{(l)} = f_{\text{GNN}} (\mathbf{A},\mathbf{H}^{(l-1)}), 
\end{equation}
where $\mathbf{A} \in \mathbb{R}^{N\times N}$ is the adjacency matrix of the graph. 
$ \mathbf{H}^{(l-1)} \in \mathbb{R}^{N \times d^{l-1}}$ denotes the node embeddings at the $(l-1)$-th GNN layer and $ \mathbf{H}^{(l)}$ is the updated node embeddings. $d^{l-1}$ is the dimension of the node embedding vectors at layer $l-1$. At the first layer, the node embeddings are usually  predefined node attributes $\mathbf{H}^{(0)} = \mathbf{X}$.

Graph convolutional networks (GCN) ~\cite{kipf2016semi} are a representative set of GNN models that have been widely studied on undirected (weighted or unweighted) graphs:
The generic framework of a GCN layer can be represented as follows:
\begin{equation}
    \mathbf{H}^{(l)} = \phi (\hat{\mathbf{A}}\mathbf{H}^{(l-1)}\mathbf{W}^{(l-1)}), 
\end{equation}
where $\hat{\mathbf{A}}$ is the normalized adjacency matrix by applying the renormalization trick to $\mathbf{A}$~\cite{kipf2016semi}.
$\mathbf{W}^{(l-1)}$ is a layer-specific trainable weight matrix, and $\phi $ is the activation function, e.g., ReLU.

Based on the GCN model, \citeauthor{deng2019learning} proposed a dynamic graph convolutional network (DynamicGCN)~\cite{deng2019learning} for forecasting protest events and identifying context graphs for the predicted events. 
This paper introduced an encoding method to encode historical news articles into a sequence of word graphs $(\mathbf{A}_{t-k+1},...,\mathbf{A}_t)$.
The adjacency matrix $\mathbf{A}_t \in \mathbb{R}^{N \times N}$ represents a word graph with $N$ nodes at time $t$ where each node is a word, and the weighted edge between two words is calculated by point-wise mutual information (PMI).
The pre-trained word embeddings are used as node features.
The proposed dynamic graph convolutional layer is defined as:
\begin{align}
    \mathbf{H}_{t} &= \phi (\hat{\mathbf{A}}_{t-1}\tilde{\mathbf{H}}_{t-1}\mathbf{W}_{t-1}), \label{equ:dygcn} 
    \quad
 \tilde{\mathbf{H}}_{t} = \left\{\begin{matrix}
 & \mathbf{H}_{0} &  \ \text{if} \ t=0\\ 
 & f([\mathbf{H}_{t} ;\mathbf{H}_{0}] ) & \text{otherwise},
\end{matrix}\right.
\end{align}
where $f(\cdot)$ is a nonlinear layer that re-encodes the input features of graph convolutional layers, including the semantic information of the words ($\mathbf{H}_0$) and the learned graph embeddings ($\mathbf{H}_t$).
The prediction is obtained by first aggregating node embeddings at the last layer to generate graph-level embeddings (e.g., linear transformation or pooling). 
The graph-level embeddings are fed to an output layer for prediction, i.e., $\hat{y}_{t^*}=f_{\text{output}}(f_{\text{agg}}(\mathbf{H}_{t}))$.
The model was demonstrated to be effective in the prediction of protest events in multiple countries. The authors also proposed a heuristic subgraph extraction to help explain event prediction results.

In addition to unstructured text data, knowledge graphs, as another data source, have been explored for temporal event prediction.
Knowledge graphs are multi-relational graphs consisting of tuples of elements (i.e., subject entity, relation, object entity), denoted as $\mathcal{G}=\{ (s, r, o)\}$. 
Nodes are entities, and edges represent relationship types. 
For example, \{(\textit{Citizen}, \underline{Criticizes}, \textit{Government}), (\textit{Citizen}, \underline{Appeals}, \textit{Police})\} can be represented as a simple knowledge graph, where \textit{Citizen}, \textit{Government} and \textit{Police} are entities and \underline{Criticizes} and \underline{Appeals} are relationship/event types.
In general, graph convolution networks for knowledge graphs can be described in terms of an update mechanism of node and edge embeddings. 
For instance, Composition-based Multi-Relational Graph Convolutional Networks (CompGCN)~\cite{vashishth2019composition} update node and edge embeddings as follows:
\begin{equation}
\mathbf{h}_o^{(l+1)}= \phi   \Big (\sum_{(s,r) \exists (s,r,o)\in \mathcal{G}} f_{\text{node}} \big ({\mathbf{h}}_s^{(l)},{\mathbf{o}}_r^{(l)} \big )  \Big ), \quad \mathbf{o}_r^{(l+1)}= f_{\text{edge}}(\mathbf{o}_r^{(l)} ),
\label{eq:compgcn-node}
 \end{equation}
 where $\mathbf{h},\mathbf{o}$ denote node and edge embeddings, respectively. 
This model jointly embeds both nodes (subjects and objects) and edges (relations) in a relational graph, and allows for more complex information modeling than GCN models.

Utilizing event-based knowledge graphs, Glean~\cite{deng2020dynamic}, was proposed to predict concurrent events of multiple types, including civil unrest events. 
This paper investigates temporal event knowledge graphs which are built upon a sequence of event sets in ascending time order $( \mathcal{G}_{t-k+1},..., \mathcal{G}_t )$, as well as a sequence word graphs $(\mathbf{A}_{t-k+1},...,\mathbf{A}_t)$.
The model includes a graph aggregation module that learns node embeddings from both knowledge graphs and semantic word graphs at each historical timestamp.
A context-aware embedding fusion module was introduced to enhance representations of nodes and edges in event knowledge graphs by blending embeddings of contextual word nodes.
The procedures can be expressed as bellows:
\begin{equation}
    \mathbf{H}_t^{\text{event}}, \mathbf{O}_t^{\text{event}} = f_{\text{CompGCN}}(\mathcal{G}_t,\mathbf{H}_{0}^{\text{event}},\mathbf{O}_{0}^{\text{event}}),  \quad 
    \mathbf{H}_t^{\text{word}} = f_{\text{GCN}}(\mathbf{A}_{t},\mathbf{H}_{0}^{\text{word}}), 
\end{equation}
\begin{equation}
    \tilde{\mathbf{H}}_t^{\text{event}} = f_{\text{Attn}} (\mathbf{H}_t^{\text{event}}, \mathbf{H}_t^{\text{word}}), \quad  \tilde{\mathbf{O}}_t^{\text{event}} = f_{\text{Attn}} (\mathbf{O}_t^{\text{event}}, \mathbf{H}_t^{\text{word}}),
\end{equation}
where $\mathbf{H}_{0}^{\text{event}},\mathbf{O}_{0}^{\text{event}}$ are randomly initialized learnable embeddings of nodes and edges in event knowledge graphs. $\mathbf{H}_{0}^{\text{word}}$ denotes node initial features of word graphs, i.e., word embeddings.
$f_{\text{Attn}}(\cdot)$ is an attention module that generates an attention vector for each node and edge in event knowledge graphs. 
Specifically, given a node in the event knowledge graph, its graph embedding is used as the context vector. The graph embeddings of the words that are semantically related to the node (obtained from $f_{\text{GCN}}(\cdot)$) are treated as the hidden states.
The attention vector of the node is obtained according to Eq. ~\ref{eq:attn}.
The enhanced representations of nodes and edges, as well as node embeddings of word graphs ($\{\tilde{\mathbf{H}}_t^{\text{event}},\tilde{\mathbf{O}}_t^{\text{event}},\mathbf{H}_t^{\text{word}}\} $) are then aggregated (e.g., pooling is applied) and fed into a GRU model for final event prediction.
This approach can also infer the potential participants of the event of interest, providing additional information to practitioners.

Interpretation is crucial in societal event prediction because it can produce supporting evidence for reliable decision-making. 
More recently, \citeauthor{deng2021understanding}~\cite{deng2021understanding} introduced 
an interpretable deep learning framework for forecasting civil unrest events and providing multilevel explanations for predictions. 
The approach aims to model three types of features including event frequencies, documents, and event graphs for a historical window, denoted as $ \{\mathbf{x},\mathcal{D},\mathcal{G}\}_{t-k+1:t}$. 
The prediction model consists of multilevel feature learning that hierarchically models heterogeneous data. It captures the dependencies between different types of data by encouraging signals to propagate from higher-level features to lower-level features:
\begin{equation}
    \mathbf{h}^{\text{freq}}_t = f_{\text{freq}}(\mathbf{x}_t),  \quad \mathbf{H}^{\text{doc}}_t = f_{\text{doc}}(\mathcal{D}_t, \mathbf{h}^{\text{freq}}_t),  \quad
    \mathbf{H}^{\text{event}}_t = f_{\text{event}} \big ( f_{\text{CompGCN}}( \mathcal{G}_t, \mathbf{H}^{\text{doc}}_t)\big ).
\end{equation}
Each type of feature is modeled at one level and then integrated to the next level.
Similar to Glean~\cite{deng2020dynamic}, it uses a recurrent neural network to model temporal information for final event prediction.

The authors also proposed an event explainer module to provide temporal and multi-level explanations for the predictor model. 
The explainer learns temporal masks for edges and node features of temporal graphs to explain the predictions via mask optimization~\cite{ying2019gnnexplainer}.

\section{Crime Prediction}\label{crime}
Crime is an unlawful event that affects the harmony of humanity. 
It can harm individuals physically and mentally, and communities that experience higher crime rates are also adversely affected.
Studying crime trends and patterns has been a top priority for law enforcement agencies, which use historical data to develop effective predictive policies to make a peaceful community~\cite{perry2013predictive}.
A wealth of research has emerged on the prediction of crime incidences.
The predictive capability can assist in crime prevention by facilitating the effective implementation of police patrols. 
In this section, we first formally introduce the problem of crime prediction, followed by a review of crime data resources.
We then summarize traditional crime prediction methods and deep learning-based approaches.

\subsection{Problem Formulation}
We first introduce some mathematical notations and the data used in this research area. Then, we formulate some widely studied crime prediction problems.

\textbf{Crime Occurrence.} Assume there are $I$ regions in an area (e.g., city) and $J$ crime categories (e.g., burglary, assault, and criminal mischief) over $T$ time slots (e.g., day, week, or month).
The region can be a longitude-latitude grid cell or a community.
We use a three-order tensor $\pmb{\mathbf{X}} \in \mathbb{R}^{I\times J \times T}$ to denote the crime data, where $x_{ijt}$ is the number of $j$-th category crime incidents committed at the $i$-th region during the $t$-th time slot. 
Given a time window of length $k$, a historical crime tensor $ \pmb{\mathbf{X}}_{t-k+1:t} = ( \mathbf{X}_{t-k+1}, ..., \mathbf{X}_t )$ is defined to represent the number of occurrences of different categories of crimes over a historical window starting from the $(t-k+1)$-th to $t$-th time slot. $\mathbf{X}_t$ is the crime matrix at time $t$, where the rows indicate the region and the columns denote the category of crime.

\textbf{External Features.} 
External features (e.g., meteorological data, traffic flow, and social media data) have been demonstrated to be tightly linked to crime incidents to facilitate the prediction of crime~\cite{wang2012spatio}. 
External features can be specified for each region $i$, each time slot $j$, or both. Simply put, we denote the external features of the whole area as $\mathcal{E}$. The external feature can be an empty set when the prediction is purely based on crime occurrence data.

\begin{definition}
\textbf{Crime Prediction as Classification}.
Given the historical crime occurrence data $\pmb{\mathbf{X}}_{t-k+1:t} $ of window size $k$ and external features $\mathcal{E}$, learn a classifier $f(\pmb{\mathbf{X}}_{t-k+1:t},\mathcal{E}) \rightarrow \mathcal{Y}_{t+\Delta}$ that infers the future crime at each region as a categorical value.
The target variable for region $j$ can be written as an integer 
$\mathcal{Y}_{t+\Delta,j} \in \{0, 1, ..., M-1 \}$.
The most common setting is a binary classification when $M$ is 2, known as crime hotspot prediction.
A hotspot is defined as a region that has a higher concentration of crime events.
In some cases, the target value may indicate the range or severity of the number of crimes of interest, usually having $M>2$.

\end{definition} 

\begin{definition}
\textbf{Crime Prediction as Regression}. Given the historical crime occurrence data $\pmb{\mathbf{X}}_{t-k+1:t} $ of window size $k$ and external features $\mathcal{E}$, the goal is to learn a regression model $f(\pmb{\mathbf{X}}_{t-k+1:t},\mathcal{E}) \rightarrow \mathcal{Y}_{t+\Delta}$ that predicts a continuous value of crime in the future time slot. 
The target variable $\mathcal{Y}_{t+\Delta}$ is usually defined as the (normalized) crime number ,  crime rate, or crime density matrix, i.e., $\mathbf{X}_{t+\Delta}$. 
\end{definition}

\begin{definition}
\textbf{Crime Prediction as Sequence Regression}. Given the historical crime occurrence data $\pmb{\mathbf{X}}_{t-k+1:t} $ of window size $k$ and external features $\mathcal{E}$, the sequence regression task is to learn a prediction model $f(\pmb{\mathbf{X}}_{t-k+1:t},\mathcal{E}) \rightarrow \{ \mathcal{Y}_{t+\Delta}, ..., \mathcal{Y}_{t+\Delta+k'-1} \}$ that predicts the crime in the next $k'$ steps. The target variables usually refer to the sequence of crime occurrences, i.e., $\{ \pmb{\mathbf{X}}_{t+\Delta}, ..., \pmb{\mathbf{X}}_{t+\Delta+k'-1} \}$.

\end{definition}

\begin{definition}
\textbf{Crime Prediction as Next-Location Prediction}. 
This problem aims to predict the location where an offender will commit a crime according to the offender’s historical trajectories or other information. Formally, given an offender $u \in \mathcal{U}$ associated with their historical crime occurrence data $\pmb{\mathbf{X}}_{t-k+1:t} $ of window size $k$, external features $\mathcal{E}$, an underlying road network $\mathcal{R}$, and historical road trajectories information $\mathcal{RT}_{t-k+1:t}^u$, the next-location prediction is to develop a model $f(\pmb{\mathbf{X}}_{t-k+1:t},\mathcal{E}, \mathcal{R}, \mathcal{RT}_{t-k+1:t}^u) \rightarrow  \mathcal{Y}_{t+\Delta}^u $ that predict the next location where an offender $u$ will commit a crime with highest likelihood.
The target variable refers to a road or a road segment in the road network.
\end{definition}
Note that crime occurrence data and target variables depend on specific studies, some of which consider each category of crime, and some combine several categories of crime into one group.

\subsection{Data Resources}
\begin{table}[]
\caption{A summary of crime data used for crime prediction. All data listed are open source. Links to these data are provided in section~\ref{sec:crime-data}.
}
\label{tab:crime-data}
\small
\scalebox{0.85}{\begin{tabular}{p{2.0cm} l p{1.4cm} p{1.5cm} p{1.4cm} p{1.4cm} p{1.0cm} p{3.4cm} p{1.8cm}}
\toprule
\textbf{Dataset}                                                & \textbf{Data level} & \textbf{Temporal coverage} & \textbf{Spatial coverage} & \textbf{Geo encoding}            & \textbf{Date encoding} & \textbf{Type}                        & \textbf{Summary}                                                                                                  & \textbf{Provided by}                     \\
\midrule
New York City (NYC) Crime Data                                    & Incident   & 2017-                                                       & New York City                                              & Coordinates             & Exact time    & All                         & Date, location, type, and other description of the crime                                                 & New York Police Department       \\
Chicago  Crime Data                                    & Incident   & 2001-2019                                                   & Chicago                                                    & Coordinates             & Exact time    & All                         & Date, location, type, and other description of the crime                                                 & Chicago Police Department       \\
San Francisco  Crime Data                              & Incident   & 2018-                                                       & San Francisco                                              & Coordinates             & Exact time    & All                         & Date, location, type, and other description of the crime                                                 & San Francisco Police Department \\
San Francisco  Crime Data (Kaggle)                     & Incident   & 2003-2015                                                   & San Francisco                                              & Coordinates             & Exact time    & All                         & Date, location, type, and other description of the crime                                                 & Kaggle                          \\
Atlanta  Crime Data                                    & Incident   & 2009-                                                       & Atlanta                                                    & Coordinates             & Exact time    & All                         & Date, location, type, and other description of the crime                                                 & Atlanta Police Department       \\
Philadelphia  Crime Data                               & Incident   & 2006-                                                       & Philadelphia                                               & Coordinates             & Exact time    & All                         & Date, location, type, and other description of the crime                                                 & Philadelphia Police Department  \\
Baltimore Crime Data                                   & Incident   & 1963-(few records in early years)                       & Baltimore                                                  & Coordinates             & Exact time    & All                         & Date, location, type, and other description of the crime                                                 & Baltimore Police Department     \\
Vancouver Crime data                                   & Incident   & 2003-                                                       & Vancouver                                                  & Coordinates             & Exact time    & All                         & Date, location, type, and other description of the crime                                                 & Vancouver Police Department     \\
Brazil Crime Data                                      & Incident   & 2007-2016                                                   & Sao Paulo                                                  & Coordinates             & Exact time    & All                         & Date, location, type, and other description of the crime                                                 & Kaggle                          \\
London Homicide Data                                   & Incident   & 2003-                                                       & London boroughs                                            & Borough                 & Month         & Homicide                    & Date, location, type, and other description of the crime                                                 & Metropolitan Police Service     \\
London Business Crime Data                             & Statistics & 2011-                                                       & London boroughs                                            & Borough                 & Month         & Bussiness crime             & Number of crimes by type, month and borough                                                              & Metropolitan Police Service     \\
London Crime Rates Data                                & Statistics & 1999-2017                                                   & London boroughs                                            & Borough                 & Year          & All                         & Numbers of crimes, and crime rates per thousand population, by type, year and borough.                   & Metropolitan Police Service     \\
London Hate Crime and Special Crime Data               & Statistics & 2012-                                                       & London boroughs                                            & Borough                 & Month         & Hate crime, special crime & Number of crimes by type, month and borough                                                              & Metropolitan Police Service     \\
London Crime Data With Different Geographic Breakdowns & Statistics & 2001-                                                       & London boroughs                                            & Borough, ward, and LSOA & Month         & All                         & Number of crimes at three different geographic levels (Borough, ward, and LSOA) by type and month        & Metropolitan Police Service     \\
Communities and Crime Dataset~\cite{asuncion2007uci}                          & /          & -1995                                                       & USA                                              & County                  & /             & /                           & A set of attributes of a county derived from socio-economic and law enforcement data, and its crime rate & UCI Machine Learning Repository \\
\bottomrule
\end{tabular}}
\end{table}

\subsubsection{Crime Data}\label{sec:crime-data}
In recent years, crime data have been published online by governmental efforts, and fine-grained crime records are available for a few big cities, such as New York~\footnote{\url{https://data.cityofnewyork.us/Public-Safety/NYC-crime/qb7u-rbmr}}, Chicago~\footnote{\url{https://data.cityofchicago.org/Public-Safety/Crimes-2019/w98m-zvie}},
San Francisco~\footnote{\url{https://data.sfgov.org/Public-Safety/Police-Department-Incident-Reports-2018-to-Present/wg3w-h783}}, Atlanta~\footnote{\url{https://www.atlantapd.org/i-want-to/crime-data-downloads}}, Philadelphia~\footnote{\url{https://www.opendataphilly.org/dataset/crime-incidents}}, Baltimore~\footnote{\url{https://data.baltimorecity.gov/datasets/part1-crime-data/explore}}, Vancouver~\footnote{\url{https://geodash.vpd.ca/opendata/}}, and London~\footnote{\url{https://data.london.gov.uk/dataset/mps-homicide-dashboard-data}}, etc.
Some are collected by online research communities, such as the crime data for Sao Paulo.~\footnote{\url{https://www.kaggle.com/inquisitivecrow/crime-data-in-brazil}}
These data are incident-based and typically contain the incident date and time, crime type, demographic details, location data, etc.
Crime can be divided into (1) violent crime such as homicide, rape, robbery, and (2) property crimes such as burglary, theft, and arson. 
In general, crime datasets in different regions follow different event definitions.
Many of the crime predictive studies are based on public datasets from the cities mentioned above~\cite{huang2018deepcrime,wang2018graph,rayhan2020aist,hu2021duronet,zhang2019analysis,hossain2020crime,balocchi2019spatial,malleson2016exploring,belesiotis2018analyzing}.
Some cities do not publish fine-grained criminal records but provide crime statistics, such as the number of crimes broken down by incident type, time, and location. 
For example, London provides statistics on homicides~\footnote{\url{https://data.london.gov.uk/dataset/mps-homicide-dashboard-data}}, business~\footnote{\url{https://data.london.gov.uk/dataset/mps-business-crime-dashboard-data}} crimes and hate crimes~\footnote{\url{https://data.london.gov.uk/dataset/recorded_crime_rates}}, and other crimes~\footnote{\url{https://data.london.gov.uk/dataset/recorded_crime_summary}}, etc.
Researchers have utilized these statistical data as ground-truth data in crime prediction~\cite{alves2018crime}. 
UCI machine learning repository~\cite{asuncion2007uci} published a crime dataset for machine learning study, Communities and Crime~\footnote{\url{https://archive.ics.uci.edu/ml/datasets/Communities+and+Crime}}, which combines socio-economic data from the 1990 US Census, law enforcement data from the 1990 US LEMAS survey, and crime data from the 1995 FBI UCR. It includes various attributes of cities in the United States and a predictor variable (i.e., crime rate).
A set of studies were conducted on this dataset~\cite{wang2012spatio,iqbal2013experimental,mcclendon2015using,anuar2015hybrid,anuar2015hybrid}. 
In addition, FBI's Crime Data Explorer (CDE) provides transparent and easy-access data in order to make awareness of criminal and noncriminal for law enforcement data sharing. It provides select datasets in the United States for download.~\footnote{\url{https://crime-data-explorer.fr.cloud.gov/pages/downloads\#nibrs-downloads}}
We organize some crime datasets in Table~\ref{tab:crime-data}.

\subsubsection{Open Source Indicators}
We introduce some external data sources that have been shown to be helpful in crime prediction.
Environmental context is often considered in crime predictive studies, such as meteorological data (e.g., temperature and weather)~\cite{wang2018graph,wang2020deep,wei2020crimestc}, demographic data (e.g., median age and race ratio)~\cite{wang2012spatio,bogomolov2014once,belesiotis2018analyzing}, geographic data (e.g., longitude and latitude)~\cite{wang2012spatio},  Point-of-Interests (POI) data (e.g., shopping, sports and education)~\cite{belesiotis2018analyzing,wang2020deep,huang2018deepcrime}, urban environmental data (e.g., noise, traffic flow, taxi trip)~\cite{belesiotis2018analyzing,wang2020deep,rayhan2020aist,wei2020crimestc}, and human behavior data (i.e., mobile data)~\cite{bogomolov2014once,zhang2019analysis,rayhan2020aist}. 
A POI is a record of a place on a map that someone finds useful or interesting, typically defined by its geographical coordinates and a few additional attributes like name and category.
Foursquare is a POI database from which some researchers have collected relevant data for crime prediction~\cite{zhang2019analysis,rayhan2020aist}. 
For human behavior-related data, some studies utilized 311 public service complaint data to assist crime prediction~\cite{duan2017deep,huang2018deepcrime}. Such data are collected from 311 Service that documents urban complaint reports of different categories from citizens through a mobile app or phone calls.
OpenStreetMap is a collaborative project that creates a free editable geographic database of the world.~\footnote{\url{https://www.openstreetmap.org/}}  In fine-grained crime analysis such as hot spot prediction, geographical information such as road network can be obtained from such data source~\cite{belesiotis2018analyzing,zhang2019analysis,ye2021spatiotemporal}.
Some researchers have explored visual data in crime prediction~\cite{zhang2019analysis}, such as Google Street View data, which provide more than 10 million miles of street view imagery across 83 countries.
Social media data have also been investigated in this line of study. For instance, Twitter posts with rich and event-based context were leveraged for predicting criminal incidents~\cite{wang2012spatio,zhang2019analysis} and next-location of crime activities~\cite{wang2015using}.

\subsection{Traditional Methods} 
Geographic information systems (GIS) were the first and most common analytical tools for spatial data. GIS is useful for mapping and retrospectively finding links between criminal structures and various spatial and social conditions~\cite{groff2002forecasting,chainey2013gis,kedia2016crime,jakobi2020gis}, but itself does not provide much predictive power.
With the increasing availability of fine-grained urban data, such as public service data, meteorological data, POI data, and human mobility data, data-driven crime prediction problems have received extensive attention from researchers for decades. 
Many data-driven approaches have emerged that enable more accurate and fine-grained predictions.

Early statistical methods for crime prediction were time series based analytical models.
\citeauthor{gorr2003short}~\cite{gorr2003short} focused on predicting crime 30 days ahead for small areas, like police precincts. In a case study of Pittsburgh, they demonstrated that simple univariate time series models including Random Walk, Brown's Simple Exponential Smoothing~\cite{brown2004smoothing}, and Holt's Two-Parameter Linear Exponential Smoothing~\cite{brown2004smoothing} are more accurate than naive methods commonly used by many police departments (i.e., Lag 12).
Several statistical time series models have been investigated in this area, such as binomial regression~\cite{wang2016crime,wang2017non} and autoregressive models.
Autoregressive Integrated Moving Average (ARIMA) was applied in one-week-ahead~\cite{chen2008forecasting} and two-year-ahead~\cite{cesario2016forecasting} crime occurrence prediction. 
Some researchers studied autoregressive models in more complex scenarios by incorporating spatial analysis~\cite{shoesmith2013space,catlett2018data,catlett2019spatio} and network analytic techniques~\cite{dash2018spatio} as well as developing hybrid models~\cite{alwee2013hybrid}.
In addition to time series models, statistical Bayesian models were developed to forecast next crime locations~\cite{liao2010novel} and to model spatial~\cite{balocchi2019spatial} or spatio-temporal~\cite{hu2018urban} patterns of urban crime.

Nonparametric approaches have also been studied.
\citeauthor{mohler2011self}~\cite{mohler2011self} implemented self-exciting point process models in the context of urban crime.  
A personalized random walk-based approach was introduced for spatial crime analysis and crime location prediction~\cite{tayebi2014crimetracer}. \citeauthor{andresen2017crime}~\cite{andresen2017crime} analyzed property crime through the use of a nonparametric spatial point pattern test that identifies the stability in spatial point patterns in pairwise and longitudinal contexts.

For traditional machine learning models, linear regression~\cite{wang2015using,mcclendon2015using,rumi2019crime,zhang2019analysis}, ridge and lasso regression~\cite{misyrlis2017spatio,belesiotis2018analyzing}, support vector regression (SVR)~\cite{alwee2013hybrid,belesiotis2018analyzing,zhang2019analysis} have been proposed to analyze the crime dynamics in regression problems.
Classification models have been extensively studied in crime hotspot prediction, including Support Vector Machine (SVM)~\cite{kianmehr2006crime,kianmehr2008effectiveness}, Logistic Regression~\cite{rumi2018crime}, Naive Bayes~\cite{iqbal2013experimental,yang2018crimetelescope}, shallow neural network~\cite{yu2011crime,rumi2018crime,yang2018crimetelescope}, k-nearest neighbors algorithm~\cite{zhang2016mixed}, clustering based model~\cite{chandra2008multivariate,yu2015hierarchical,ingilevich2018crime}, decision tree~\cite{almanie2015crime,saltos2017exploration}, and ensemble models~\cite{iqbal2013experimental,bogomolov2014once,baculo2017geospatial,alves2018crime,yang2018crimetelescope,rummens2017use,rumi2018crime,yu2011crime,hossain2020crime,lamari2020predicting}.
Some researchers studied more advanced approaches.
\citeauthor{zhao2017modeling}~\cite{zhao2017modeling} modeled temporal-spatial correlations into a coherent optimization framework for crime prediction.
\citeauthor{kadar2019public}~\cite{kadar2019public} developed an imbalance-aware hyper-ensemble for spatio-temporal crime prediction, focusing on low population density areas.
\citeauthor{yi2018integrated}~\cite{yi2018integrated} proposed a Clustered Continuous Conditional Random Field (Clustered-CCRF) model that exploits both spatial and temporal factors for crime prediction in an integrated way.

Most traditional methods rely on feature engineering and have limitations when modeling complex crime data. Recently, deep learning models have proven to be effective in crime prediction. We introduce the deep learning-based approach in the next section.

\subsection{Deep Learning Models}
\label{sec:crime-deep}

\begin{table}[]
\caption{Deep learning based crime prediction techniques. Note that the intensity or rate of crime derives from the number of crimes, so we do not distinguish them in the inference column. 
We use horizontal lines to partition the models with different structural characteristics, corresponding to section~\ref{sec:crime-deep}.
}
\footnotesize
\label{tab:crime-models}
\begin{threeparttable}
\scalebox{0.85}{\begin{tabular}{p{1.8cm} p{1.6cm} p{1.2cm} p{1.6cm} p{2.0cm} p{1.6cm} p{1.4cm} p{1.4cm} p{1.6cm} p{2.0cm}}
\toprule
\textbf{Technique}                         & \textbf{Study area}                                                          & \textbf{Time}                                       & \multicolumn{2}{l}{\textbf{Dataset}}                                                                       & \multicolumn{4}{l}{\textbf{Forecasting}}                                                                                & \textbf{Evaluation}                                       \\
\cmidrule(r){4-5}\cmidrule(r){6-9}
                                  &                                                                     &                                                       & \textbf{Crime type}                   & \textbf{External data}                                                      & \textbf{Inference}                  & \textbf{Task}                & \textbf{Temporal unit}   & \textbf{Spatial unit}                              &                                                  \\

\midrule
FFNN~\cite{chun2019crime}                 & /                                                                   & 1997-2017                                             & /                            & /                                                                  & Severity of personal crime & Multi-label class.              & 5 years         & /                                         & Accuracy, F1                                     \\
FFNN~\cite{kang2017prediction}            & Chicago                                                             & 2013-2014                                             & All crime                    & Demographic, housing, economic, education, weather, and image data & Hotspots                   & Binary class.       & Day             & 0.001 latitude$\times$0.001 longitude grid cells & Accuracy, Precision, Recall, AUC                 \\
STCN~\cite{duan2017deep}                  & NYC                                                                 & 2010-2015                                             & Felony                       & 311 data                                                           & Hotspots                   & Binary class.       & Day             & Grid cells of 0.18$km^2$                    & F1, AUC                                          \\
ST-ResNet~\cite{wang2019deep}            & Los Angeles                                                         & 2015                                                  & All crime                    & Weather, holiday  & \# of crimes               & Reg.                & Hour            & Grid cells                                & RMSE, Top-N-Accuracy                             \\
DIRNet~\cite{ye2021spatiotemporal}      & NYC                                                                 & 2010-2015                                             & 3 theft crime types  & 311 data                                                           & Hotspots                   & Binary class.       & Day             & Grid cells of 0.28$km^2$                    & F1                                               \\
DuroNet~\cite{hu2021duronet}             & Chicago, NYC                                                        & 2016-2017, 2015-2016                                  & All crime                    & /                                                                  & \# of crimes               & Reg.                & Day             & Communities, Police precincts             & RMSE, MAE                                        \\
\midrule
Sacked LSTM~\cite{wawrzyniak2018data}   & Poland                                                              & 2008-2014, 2013-2016                                  & 12 crime types               & /                                                                  & \# of crimes               & Reg.                & Day, Week, Year & Regions                                   & MSE                                              \\
STNN~\cite{zhuang2017crime}              & Portland, Oregon                                                    & 2012-2016                                             & 3 crime types                & /                                                                  & Hotspots                   & Binary class.       & 2 weeks         & Grid cells of 0.34$km^2$                    & Accuracy, Precision, Recall, F1                  \\
RCN~\cite{stec2018forecasting}           & Chicago, Portland                                                   & 2001-2018, 2012-2017                                  & All crime                    & Weather, public transportation, and census data                    & \# of crimes (in bins)              & Class.              & Day             & Police beats, grid cells                  & MASE, Accuracy                                   \\
CNN and RNN~\cite{stalidis2018examining} & Seattle, Minneapolis, Philadelphia, San Fransisco, Washington, D.C. & 1996-2016, 2010-2016, 2006-2017, 2003-2015, 2008-2017 & All crime                    & /                                                                  & Hotspots, \# of crimes     & Binary class., Reg. & Day             & Grid cells of 0.2$km^2$  and 0.64$km^2$            & F1, AUROC, AUPR, PAI  \\
GSRNN~\cite{wang2018graph}               & Chicago, Los Angeles                                                & 2015, 2014-2015                                       & All crime                    & /                                                                  & \# of crimes               & Reg.                & Hour            & Zip code regions                          & RMSE, Precision Matrix                           \\
\midrule
DeepCrime~\cite{huang2018deepcrime}      & NYC                                                                 & 2014                                                  & 4 crime types                & POIs and 311 data                                                 & Hotspots                   & Binary class.              & Day             & Police precincts                          & F1                                               \\
MiST~\cite{huang2019mist}                & NYC, Chicago                                                        & 2015                                                  & 4 crime types, 8 crime types & /                                                                  & Hotspots                   & Binary class.              & Day             & 2$km$×2$km$  grid cells                      & F1, AUC                                          \\
\midrule
NN-CCRF~\cite{yi2019neural}             & NYC, Chicago                                                        & 2015-2016, 2013-2015                                  & 2 crime types                & /                                                                  & \# of crimes               & Reg.                & Day             & 1km×1km grid cells                        & RMSE, Hits@k                                     \\
Crime-GAN~\cite{jin2019crime}            & San Francisco                                                       & 2003-2018                                             & 4 crime types                & /                                                                  & \# of crimes               & Sequence reg.       & Week            & Grid cells                                & RMSE, MSPE, JS      \\
CSAN~\cite{wang2020csan}                 & San Francisco                                                       & 2003-2018                                             & 4 crime types                & /                                                                  & \# of crimes               & Sequence reg.       & Week            & Grid cells                                & RMSE, MSPE, JS                                  \\
\midrule
TGCN~\cite{jin2020addressing}            & San Francisco                                                       & 2003-2019                                             & 4 crime types                & /                                                                  & \# of crimes               & Sequence reg.       & Week            & Grid cells                                & RMSE, MSPE, JS                                  \\
DT-MGCN~\cite{wang2020deep}              & Chicago                                                             & 2001-2014                                             & 10 crime types               & Census, Taxi flow, traffic violations, POIs data                   & \# of crimes               & Sequence reg.       & Month           & Communities                               & MAE, RMSE, MRE             \\
CrimeSTC~\cite{wei2020crimestc}          & NYC                                                                 & 2014                                                  & 4 crime types                & Meteorology, urban anomaly, demographic and POIs data              & \# of crimes               & Reg.                & Day             & Police precincts                          & RMSE, MAE                                        \\
ST-SHN~\cite{xiaspatial}                 & NYC, Chicago                                                        & 2014-2015, 2016-2017                                  & 4 crime types                & /                                                                  & Hotspots, \# of crimes     & Binary class., Reg. & Day             & 3$km \times$3$km$  grid cells                        & F1, RMSE, MAE                       \\  
\bottomrule
\end{tabular}}
\begin{tablenotes}\footnotesize
\item [Note]mAP (Mean average precision); PAI (Prediction Accuracy Index); JS (Jensen-Shannon Divergence); MRE (Mean Relative Error)
\end{tablenotes}
\end{threeparttable}
\end{table}

Criminal events are often closely related to time and space, showing complex reoccurring patterns. Researchers have studied various deep learning-based models to capture spatial and temporal dependencies to predict crime and have made considerable progress.
We present existing work according to model characteristics, mainly in the categories of Feedforward Neural Network (FNN), Recurrent Neural Networks (RNN), Attention, Autoencoder and Deep Generative approach, and Graph Neural Network (GNN).
We provide a summary of deep learning-based models for crime prediction in Table~\ref{tab:crime-models}. 

\subsubsection{Feedforward Neural Network Based Approaches} 
A feed-forward neural network (FNN) is a network without recurrent connections.
One typical network is the fully-connected feed-forward neural network (FFNN), also known as multi-layered perceptron (MLP).
\citeauthor{chun2019crime}~\cite{chun2019crime} applied an FFNN model on individuals' criminal charge history to predict the severity of the crime at the individual level over the next 5 years.
\citeauthor{kang2017prediction}~\cite{kang2017prediction} introduced an FFNN based model with a feature-level data fusion method for predicting daily crime hotspots.
The model consists of several feature layers for multi-level feature representation and abstraction, where each feature layer operates independently using a set of features.
They are then concatenated for joint feature representation learning.
The authors introduced three sets of features, including spatial, temporal features, and environmental context features extracted by a CNN using image data.

\begin{figure}[h]
  \centering
  \begin{minipage}[b]{0.38\textwidth}
    \includegraphics[width=\textwidth]{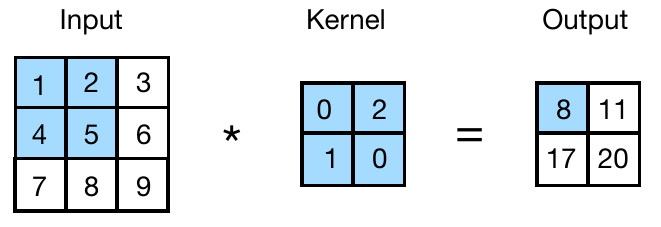}
    \caption{Two-dimensional convolution operation with a kernel of size $2 \times 2$.}
    \label{fig:2d-conv}
  \end{minipage}
\hspace{60pt}
  \begin{minipage}[b]{0.32\textwidth}
    \includegraphics[width=\textwidth]{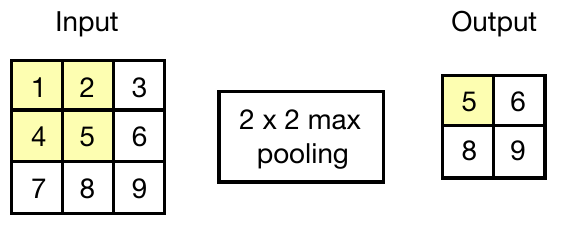}
    \caption{Maximum pooling with a pooling window of size $2 \times 2$.}
    \label{fig:pool}
  \end{minipage}
\end{figure}

CNN is a class of feedforward neural networks that can extract features from data with convolution kernels/filters.
CNN has been extensively studied in crime prediction and has shown advantages in improving prediction accuracy by learning spatial features in crime data. 
The convolutional and pooling layers are two main building blocks of CNNs.
Fig.~\ref{fig:2d-conv} shows a simple example of convolution.
In this example, the input is a two-dimensional tensor of size $3\times 3$, and the kernel/convolution window is a two-dimensional tensor of size $2\times 2$.
The convolution window slides over the input tensor and performs element-wise multiplication with corresponding elements. 
The resulting tensor is summed up yielding a single scalar value.
Convolution operation usually involves several other techniques such as padding, stride, and dilation, which control the size of the output. 
Pooling is a downsampling method that reduces the spatial size of the representation as well as the number of parameters and computation in the network. Fig.~\ref{fig:pool} shows an example of maximum pooling.

\begin{figure}[h]
  \centering
  \begin{minipage}[b]{0.46\textwidth}
    \includegraphics[width=\textwidth]{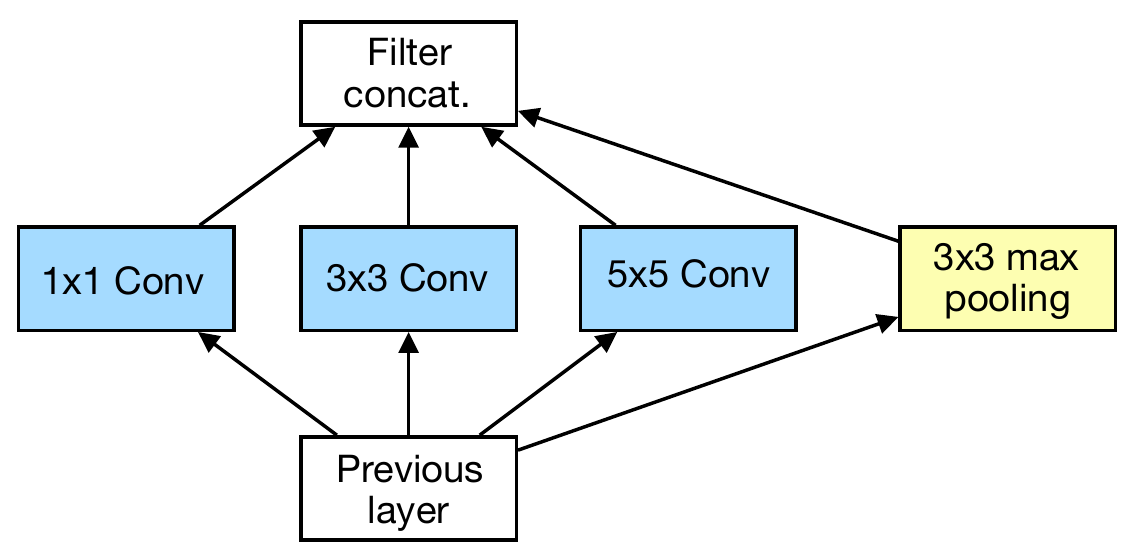}
    \caption{Inception module, naive version~\cite{szegedy2016rethinking}.}
    \label{fig:inception}
  \end{minipage}
\hspace{40pt}
  \begin{minipage}[b]{0.3\textwidth}
    \includegraphics[width=\textwidth]{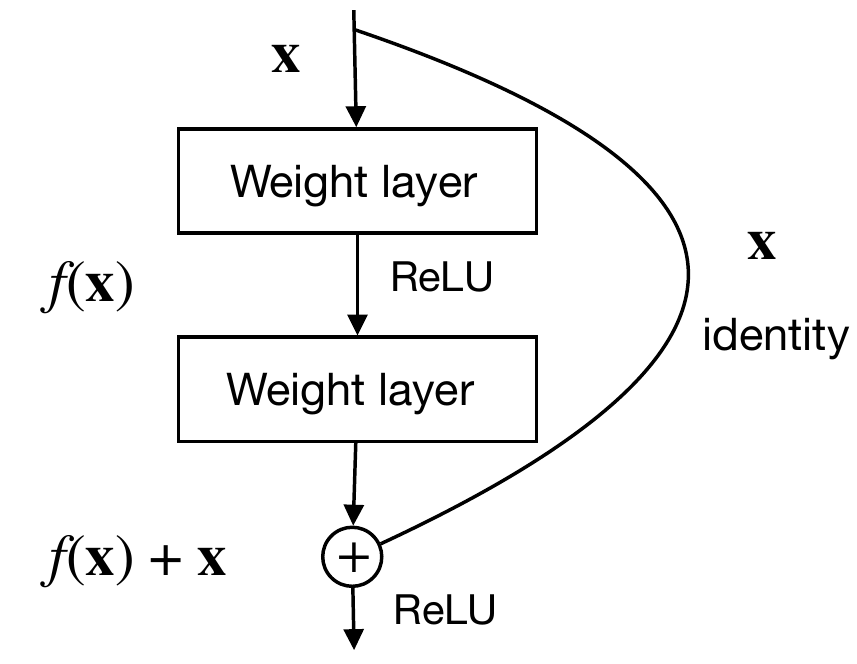}
    \caption{A residual block~\cite{he2016deep}.}
    \label{fig:residual}
  \end{minipage}
\end{figure}

\citeauthor{fu2018streetnet}~\cite{fu2018streetnet} developed a CNN model with stacked convolutional layers to extract latent features of street view images to infer the ranking of crime types. 
Instead of stacking convolution layers deeper and deeper in a CNN model, many effective and computationally efficient architectures have been developed. 
Some of them were adopted for crime prediction, such as Inception Network~\cite{szegedy2016rethinking} and Residual Neural Network (ResNet)~\cite{he2016deep}. 
An inception network is a deep neural network with an architectural design that consists of repeating components referred to as inception modules, as shown in Fig.~\ref{fig:inception}.
A residual network, as shown in Fig.~\ref{fig:residual}, consists of residual blocks which have skip connections, also called identity connections. The identity mapping helps in tackling the vanishing gradient problem in networks with a large number (even thousands) of layers.

Based on the achievements of CNN models,  \citeauthor{duan2017deep} proposed a CNN-based Spatiotemporal Crime Network (STCN)~\cite{duan2017deep}, to predict the occurrence of felonies on the next day. 
They employed crime event counts and 311 data and converted them into two 2D image-like arrays as input features.
The STCN consists of convolutional layers, an inception block, a fractal block~\cite{larsson2016fractalnet} (i.e., an alternative of the residual block that can achieve excellent performance as standard residual networks), pooling layers, and fully connected layers. 
Convolutional layers capture low-level spatio-temporal dependencies of criminal events, and the inception and fractal blocks intend to abstract high-level spatio-temporal features.
\citeauthor{wang2019deep}~\cite{wang2019deep} applied a popular deep CNN architecture, Spatio-Temporal Residual Networks (ST-ResNet)~\cite{zhang2017deep}, to predict crime distributions over the Los Angeles area on an hourly timescale. 
The ST-ResNet captures crime dynamics through convolutional layers and employs residual neural networks to model the temporal features of crimes. 
Due to the low regularity of crime data in both space and time, the authors preprocessed the crime data to select appropriate spatio-temporal scales for prediction and proposed different data regularization methods for spatial and temporal dimensions.
Furthermore, to address the problem of the sparsity of crime events in space and time, 
~\citeauthor{ye2021spatiotemporal}~\cite{ye2021spatiotemporal} proposed a deep inception-residual network (DIRNet) to conduct theft-related crime prediction based on crime and 311 data.
This framework consists of convolutional layers and inception layers to extract spatiotemporal dependencies from crime and 311 data, respectively. 
The learned features are then combined and fed to stacked residual layers to capture high-level feature interactions for final prediction.

\subsubsection{Recurrent Neural Network Based Approaches}
RNNs have been developed to model the temporal dependencies of time-series crime data.
In the context of crime prediction, \citeauthor{wawrzyniak2018data}~\cite{wawrzyniak2018data} studied short-term prediction for a selected crime type at daily, weekly, and annually levels by using a stacked LSTM architecture.
The model does not handle spatial information and performs prediction for each region separately.
Since crime is not randomly distributed geographically, considering the status of neighboring regions is beneficial for predicting the target region.
\citeauthor{zhuang2017crime}~\cite{zhuang2017crime} formulated the problem of crime forecasting as a space-time series prediction problem and introduced a Spatio-Temporal neural network (STNN), where RNN models are fed with spatial embeddings. The spatial embeddings are the accumulation of the short-term crime numbers of a space-time window.

Some researchers combine CNN and RNN models to handle both the temporal and spatial features of crime patterns.
\citeauthor{stec2018forecasting}~\cite{stec2018forecasting} used a joint recurrent and convolutional neural network (RCN) to predict crime while combining crime data with additional weather, public transportation, and census data. 
In this network, spatial features are passed through the CNN layer, and the output is combined with other features and then fed to the RNN model.
Another work investigated the capability of several deep learning architectures, including CNNs and RNNs, to forecast crime hotspots in an urban environment~\cite{stalidis2018examining}. This work examined three deep learning architecture configurations for crime prediction based on the encoding order of spatial and temporal patterns, where spatial features are modeled using CNN models such as ResNet, and RNNs learn temporal features.

\begin{figure}[h]
    \centering
    \includegraphics[width=.5\linewidth]{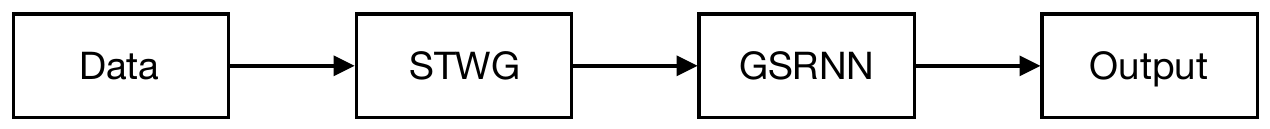}
    \caption{The flow chart of the GSRNN framework.~\cite{wang2018graph}.}
    \label{fig:gsrnn}
\end{figure}

Researchers have begun to investigate diverse representations of data for crime analysis.
To address the challenges of forecasting spatio-temporal distributions at hourly or even finer temporal scales for crime data, 
\citeauthor{wang2018graph}~\cite{wang2018graph}  developed a graph-based multiscale framework to model sparse and unstructured spatial-temporal data.
The flow chart of the framework is shown in Fig.~\ref{fig:gsrnn}.
The framework contains two components. The first part is spatial-temporal weighted graph (STWG) inference. It results in a graph representation for the spatio-temporal evolution of the data. 
Each graph node is associated with a time series of crime intensity in a region, and the graph topology is inferred from the self-exciting point process model. 
The graph representation captures versatile spatial partitioning that enables forecasting at different spatial scales. 
The second component is a deep learning model that approximates the temporal evolution of the data. They introduced a graph-structured RNN (GSRNN) on the inferred graph based on the structural-RNN (SRNN)~\cite{jain2016structural} architecture. 
The GSRNN arranges RNNs in a feedforward manner.
It first assigns a cascaded LSTM to fit the time series on each node in the graph. Simultaneously, it associates each edge in the graph with a cascaded LSTM that receives the output from neighboring nodes along with the weights learned from the Hawkes process.
The specially designed deep neural network takes advantage of the RNN’s ability to learn time series patterns, capturing real-time interactions of each node to its connected neighbors. 

\subsubsection{Attention Based Approaches}
Many studies have shown that the combination of attention mechanisms in RNNs allows them to focus on parts of the input sequence when predicting the output sequence, making learning easier and of higher quality.
With the broad exploration of attention in various research fields, more advanced attention mechanisms have been studied to model complex and rich features and improve the predictive ability of crime prediction.
\begin{figure}[h]
    \centering
    \includegraphics[width=.85\linewidth]{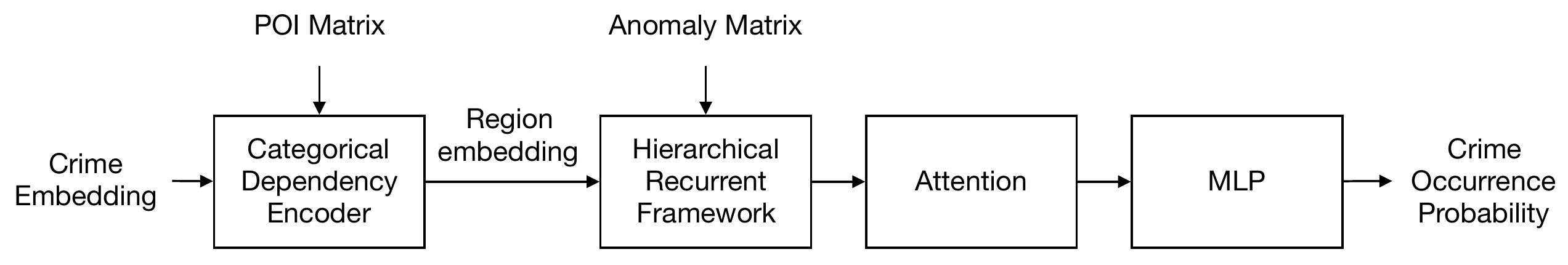}
    \caption{The DeepCrime Framework~\cite{huang2018deepcrime}.}
    \label{fig:deepcrime}
\end{figure}
In this regard, \citeauthor{huang2018deepcrime} proposed an attentive hierarchical recurrent network, DeepCrime~\cite{huang2018deepcrime}, that uncovers dynamic crime patterns and explores the evolving inter-dependencies between crimes and other ubiquitous data in urban space.
The framework, as shown in Fig.~\ref{fig:deepcrime}, first models region-category interactions to generate a region embedding vector.
It then includes a three-level GRU architecture that encodes the temporal dynamics of crime patterns and their interrelations with urban anomalies (i.e., 311 anomalies).
The model further consists of an attention layer to capture the unknown temporal relevance and automatically assign importance
weights to the learned hidden states at different time frames.
The MLP is then applied to the learned attention features to output crime predictions.
\begin{figure}[H]
    \centering
    \includegraphics[width=.89\linewidth]{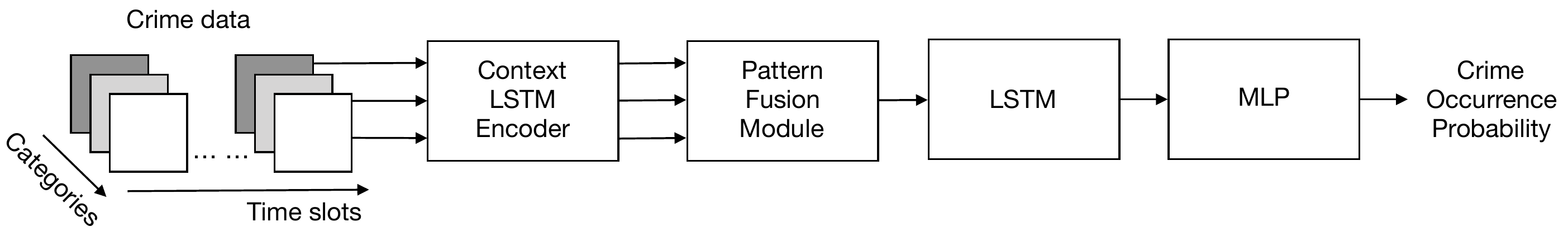}
    \caption{The MiST Framework~\cite{huang2019mist}.}
    \label{fig:huang2019mist}
\end{figure}
More recently, \citeauthor{huang2019mist} studied a Multi-View and Multi-Modal Spatial-Temporal learning framework (MiST)~\cite{huang2019mist}, to explicitly model the dynamic patterns of citywide abnormal events from spatial-temporal-categorical views. 
Fig.~\ref{fig:huang2019mist} presents the framework of the MiST model.
The framework first models dynamic intra-region correlations from temporal views with a context-LSTM encoder. Then, a pattern fusion module with the attention mechanism captures complex inter-region and cross-categorical correlations from spatial-categorical views. Additionally, it incorporates an LSTM to capture the sequential patterns of cross-modal correlations between locations, times, and event categories. 
The proposed approach was evaluated on real-world crime and urban anomaly data and compared to state-of-the-art baselines to demonstrate its superior performance.

Instead of combining recurrent neural networks with attention, ~\citeauthor{hu2021duronet} incorporated attention into convolutional networks.
They studied the effect of noises (i.e., local outliers and irregular waves) on crime data and proposed a Dual-robust Enhanced Spatial-temporal Learning Network (DuroNet)~\cite{hu2021duronet} with an encoder-decoder architecture to capture deep crime patterns.
The encoder consists of (1) a CNN based locality enhanced module that employs local temporal context information to smooth the deviation of outliers and uses gate mechanism to enhance spatial-temporal representation, and (2) a self-attention based pattern representation module to weaken the effect of irregular waves by learning attentive weights. 
A self-attention module compares every element in the sequence to every other element in the sentence, including itself, and reweigh the embeddings of each element to include contextual relevance.
Finally, a feed-forward prediction network with convolutional layers serves as the decoder for crime prediction.

\subsubsection{Autoencoder and Deep Generative Approaches} 
\begin{figure}[h]
    \centering
    \includegraphics[width=.5\linewidth]{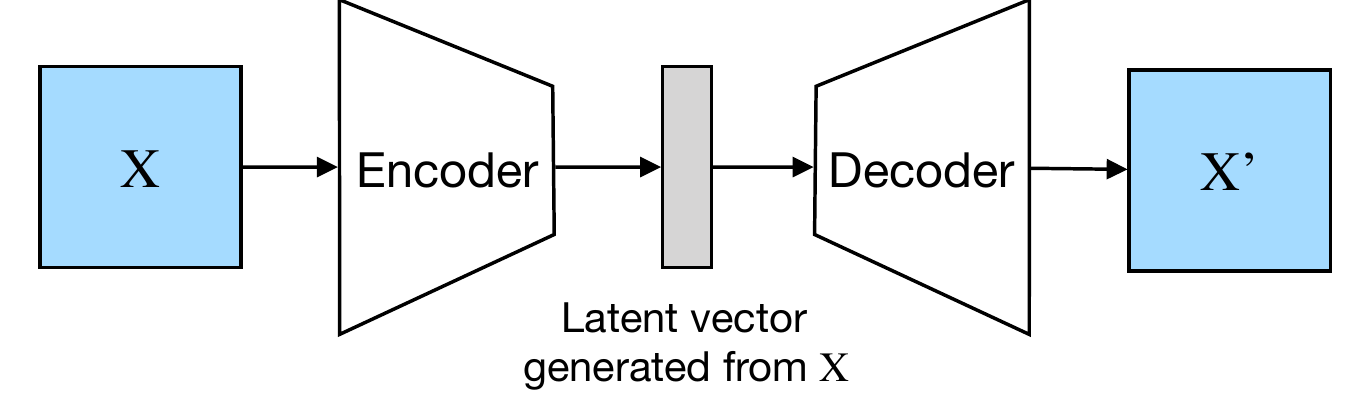}
    \caption{ Illustration of the autoencoder model architecture.}
    \label{fig:autoencoder}
\end{figure}

Autoencoders are a type of self-supervised learning model that can learn a compressed/reconstructed representation of input data, which are commonly used for feature selection and extraction. An autoencoder is composed of an encoder and a decoder sub-models, as shown in Fig.~\ref{fig:autoencoder}.

\citeauthor{yi2019neural}~\cite{yi2019neural} adopted autoencoder to learn dynamic spatial correlations in a deep learning framework. To achieve fine-grained crime prediction at the daily level, this paper
leveraged the mean-field approximation theory to simplify the inference of continuous conditional random field (CCRF)~\cite{ristovski2013continuous} model.
They proposed an end-to-end neural network NN-CCRF\cite{yi2019neural} to model temporal and spatial correlations, respectively. 
The traditional CCRF model is one of the probabilistic graphical models that can explore sequential relationships in time series data.
It consists of two parts, unary potential, and pairwise potential.
The NN-CCRF model applies LSTM as the unary potential and leverages Stacked Denoising AutoEncoder (SDAE) to learn spatial correlations across regions used in pairwise potential.
The Denoising AutoEncoder~\cite{vincent2010stacked} attempts to obtain robust latent representations by introducing a stochastic noise to the original data.

\begin{figure}[h]
    \centering
    \includegraphics[width=.32\linewidth]{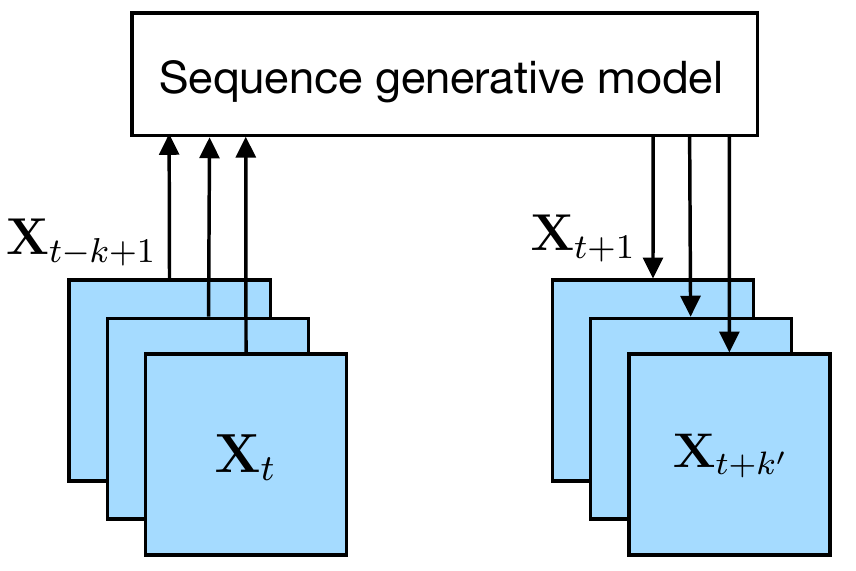}
    \caption{Illustration of the crime sequence prediction task. A sequence generative model generates the next $k'$ crime distributions conditioned on a sequence of previous $k$ crime distributions.}
    \label{fig:seq-gen}
\end{figure}

Recently, sequence generative networks have gained considerable attention in various fields such as computer vision and natural language processing. 
Researchers have found it promising in capturing the spatio-temporal dynamics of massive data. Some studies provided new insights into understanding complex phenomena in crime data from sequence generation.
\citeauthor{wang2020csan} proposed a sequence generative neural network, named Crime Situation Awareness Network (CSAN)~\cite{wang2020csan}, for crime sequence prediction.
The task is illustrated in Fig.~\ref{fig:seq-gen}. 
The CSAN includes two components: (1) a compact representation model with Multiple-VAEs attempts to embed high-dimensional sparse crime data into some compact latent spaces; (2) a crime situation generation model with GRU and attention layers captures potential dynamics in latent space, leading to sequence prediction.
Variational Autoencoder (VAE)~\cite{kingma2013auto} consists of an encoder that compresses the input data into a constrained multivariate latent distribution, and a decoder that reconstructs the data given the latent distribution.

\begin{figure}[]
    \centering
    \includegraphics[width=.56\linewidth]{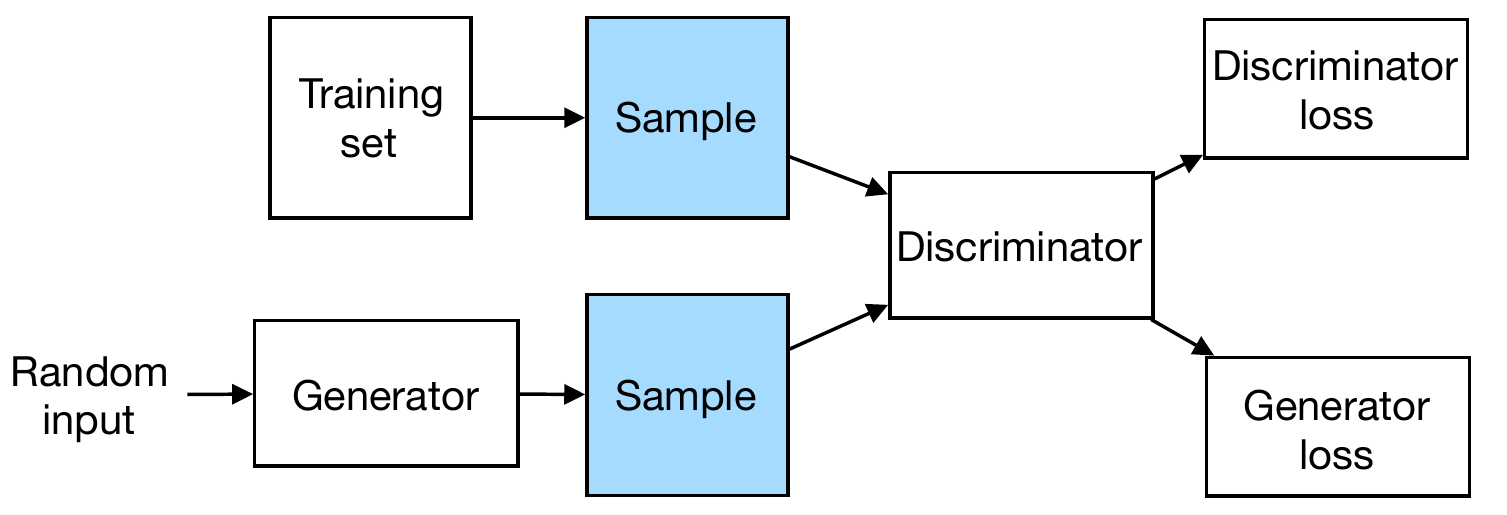}
    \caption{The framework of the generative adversarial network (GAN) model.}
    \label{fig:gan}
\end{figure}

There is another effort focusing on crime sequence generation.
\citeauthor{jin2019crime}  developed a context-based sequence generative model in a generative adversarial network, Crime-GAN~\cite{jin2019crime}.
Generative adversarial network (GAN)~\cite{NIPS2014_5ca3e9b1} is a deep generative model that has shown tremendous success over the last few years.
GAN is built on the principle of adversarial loss. It includes a Generator and a Discriminator, which follows adversarial nature to fool each other. The framework is presented in Fig.~\ref{fig:gan}.
In the Crime-GAN model, a CNN-based VAE first extracts latent variables from the crime data. Then, the latent variables are fed into a GRU-based Seq2Seq model to generate latent variables for the next $k'$ time step. The model uses the GRU model as a discriminator to examine whether the generated sequence of crime latent variables follows the previous sequence of crime latent variables.

\subsubsection{Graph Neural Network Based Approaches}  
Graph neural networks have shown the ability to capture spatially structured features beyond the Euclidean distance. There have been attempts to incorporate such networks in crime prediction.
\citeauthor{jin2020addressing} proposed a graph deep learning approach, Temporal Graph Convolutional Network (TGCN)~\cite{jin2020addressing} to predict crime sequences.
The authors first constructed a sequence of crime spatial correlation graphs using the Pearson coefficient. Each node in the graph represents a region.
TGCN consists of three main components: (1) a GCN~\cite{kipf2016semi} to extract latent representations from a defined crime spatial correlation graph; (2) an LSTM to capture temporal dynamics of the latent space; and (3) a CNN-based model to output crime predictions.
Researchers have incorporated more sophisticated graph information in crime sequence prediction.
\citeauthor{wang2020deep} proposed a Deep Temporal Multi-Graph Convolutional Network (DT-MGCN)~\cite{wang2020deep} that integrates a graph generation component and a spatial-temporal component to forecast spatial-temporal crime rate.
The authors encoded various external factors into multiple graphs, including distance graph, interaction graph, and correlation graph, to capture the Euclidean and non-Euclidean correlations among communities, and generated a fused graph by combining the aforementioned graphs.
The spatial-temporal component simultaneously employs spectral graph convolutions to capture spatial patterns and uses an encoder-decoder temporal convolutional network (EDTCN)~\cite{lea2017temporal} to model temporal features.

\begin{figure}[h]
    \centering
    \includegraphics[width=.81\linewidth]{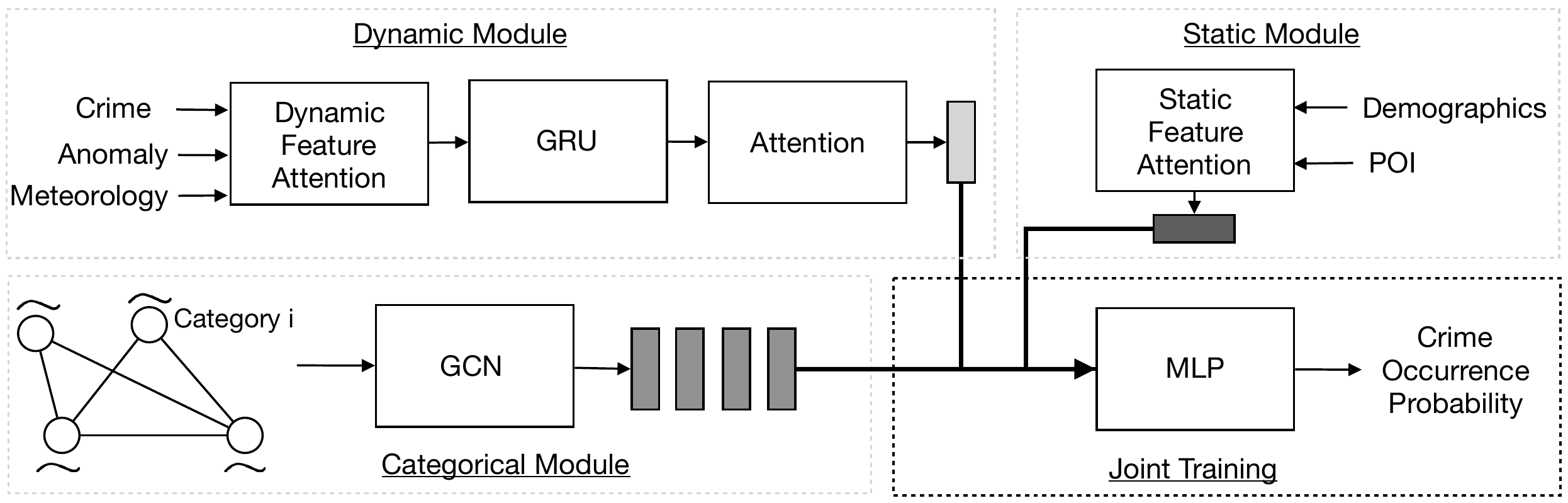}
    \caption{The CrimeSTC Framework~\cite{wei2020crimestc}.}
    \label{fig:wei2020crimestc}
\end{figure}
\begin{figure}[h]
    \centering
    \includegraphics[width=.8\linewidth]{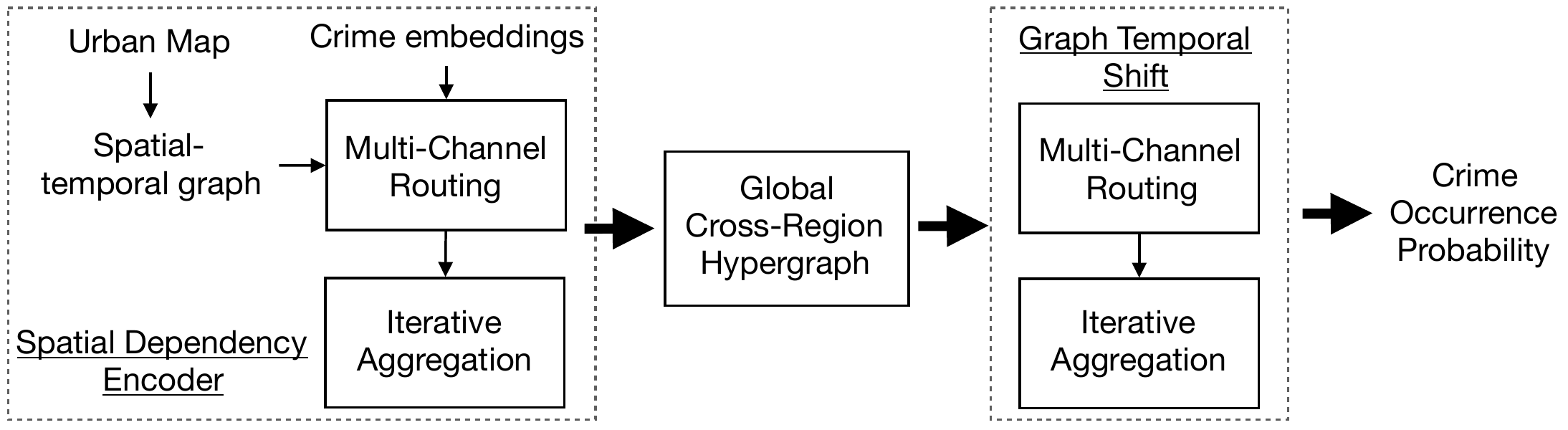}
    \caption{The ST-SHN Framework~\cite{xiaspatial}.}
    \label{fig:stshn}
\end{figure}

Graph neural networks have also been used to capture the dependencies among different crime categories to help predict the number of crimes.
In particular, \citeauthor{wei2020crimestc} proposed a crime prediction framework, CrimeSTC~\cite{wei2020crimestc} to jointly learn intricate spatial-temporal categorical correlations hidden inside crime and big urban data.
The model framework is shown in Fig.~\ref{fig:wei2020crimestc}. 
The model consists of four parts: (1) a dynamic module that processes temporal data through local CNN, GRU, and attention; (2) a static module handling static data through fully connected layers; (3) a categorical module that captures categorical dependencies through the GCN model; and (4) a joint training module concatenating dynamic and static representations to forecast crime numbers.
More recently, \citeauthor{xiaspatial} proposed a Spatial-Temporal Sequential Hypergraph Network (ST-SHN)~\cite{xiaspatial} architecture that achieves the state-of-the-art performance in crime occurrence prediction.
ST-SHN collectively encodes complex crime spatial-temporal patterns as well as the underlying category-wise crime semantic relationships. The model framework is provided in Fig.~\ref{fig:stshn}.
Specifically, the authors proposed an attention-based multi-channel routing mechanism to learn cross-type crime influences under the graph neural network framework. They designed a graph message passing architecture that integrates a hypergraph learning paradigm to enhance the cross-region relation learning without the limitation of adjacent connections.
To model the temporal dependencies of crime, a graph temporal shift mechanism is developed to inject the time-varying spatial-temporal crime patterns into the representation process.
 
\section{Deep Learning Techniques}\label{tech}
Deep learning is a family of machine learning methods based on artificial neural networks and representation learning.
In this section, we summarize the utilization of deep neural networks in civil unrest and crime prediction.

\paragraph{Convolutional Neural Networks} 
CNN is designed to automatically learn spatial hierarchies of features through backpropagation by using multiple building blocks, such as convolution layers, pooling layers.
Compared to limited research on CNNs in civil unrest prediction 
more studies have explored CNNs in crime prediction~\cite{zhang2019analysis,duan2017deep,wang2019deep,ye2021spatiotemporal,hu2021duronet}.
This is mainly because that the spatial granularity of criminal events is usually smaller than civil unrest.
Civil unrest such as demonstrations and riots can occur throughout a city.
It usually produces a massive impact, spreading online or offline to influence people's behavior in surrounding areas or other cities.
In contrast, crimes such as burglary and robbery may be more specific to a street or community.
In some studies that focus on fine-grained crime prediction, researchers divide a region (e.g., a city) into grid cells and forecast crime for each cell.
CNN provides an adaptive and effective way to learn spatial correlations among cells.

\paragraph{Recurrent Neural Networks} 
A recurrent neural network (RNN) is a class of artificial neural networks with recurrent connections. It uses the internal state (memory) to process input sequences of various lengths.
Such models have been widely involved in societal event prediction and often outperform statistical time series models such as autoregressive models.
RNN based methods capture nonlinearities and more complex dependencies in sequential data.
In dynamic social environments, societal events usually do not occur independently. They are are often influenced by past events.
This explains why historical time series data are used as important indicators when forecasting civil unrest and crime.
To better capture temporal dependencies, researchers have used RNNs and their variants (i.e., LSTM, GRU) as the main components in temporal feature learning for civil unrest~\cite{smith2017predicting,parrish2018crystal,halkia2020conflict,mengleveraging,wang2018unrest} and crime~\cite{wawrzyniak2018data,zhuang2017crime,stec2018forecasting,stalidis2018examining,wang2018graph,huang2018deepcrime,huang2019mist,yi2019neural} prediction.

\paragraph{Attention}
Attention models, or attention mechanisms, are techniques in neural networks that allow a network to focus on specific aspects of a complex input, guided by a context vector (i.e., query).
Attention is usually applied together with RNN models and is able to show the importance of features through attention weights.
In the context of civil unrest prediction, 
\citeauthor{wang2018unrest}~\cite{wang2018unrest} introduced the attention mechanism to discover crucial historical data points.
\citeauthor{ertugrul2019activism}~\cite{ertugrul2019activism} studied a more advanced attention-based approach, where attention mechanisms are used to explain the importance of activities across different historical time steps and different regions.
Some researchers applied attention to enhance representation learning with contextual information~\cite{deng2020dynamic}. 
With similar motivations, more sophisticated models have introduced attention mechanisms in crime prediction~\cite{huang2018deepcrime,huang2019mist,wei2020crimestc}.

\paragraph{Autoencoder and Deep Generative Approach} 
Autoencoder is a type of artificial neural network used to learn efficient data representations in an unsupervised manner.
\citeauthor{yi2019neural}~\cite{vincent2010stacked} leveraged the Denoising AutoEncoder to learn latent representations to capture spatial correlations across regions for crime prediction.
Others utilized Variational AutoEncoder (VAE) to learn a compressed latent representation of crime matrices~\cite{wang2020csan,jin2019crime}.
VAE can be regarded as a deep generative model.
Deep generative models provide a powerful way to approximate complicated and high-dimensional probability distributions using a large number of samples.
Some researchers have incorporated generative adversarial networks (GAN) to estimate future crime distributions across a region in a sequential predict task~\cite{jin2019crime}.

\paragraph{Graph Neural Networks}
Graph Neural Networks (GNNs) are a class of deep neural networks that can be directly applied to graphs and provide an easy way for node-level, edge-level, and graph-level prediction tasks. 
In civil unrest and crime prediction, researchers have explored network structures for different purposes. \citeauthor{deng2019learning}~\cite{deng2019learning} proposed a dynamic graph neural network to predict civil unrest, which takes a sequence of dynamic word graphs as input. They further discovered key contextual subgraphs as supporting evidence for event prediction. The subgraphs provide structured and concise data representations which can streamline the process of event analysis.
Other studies have applied GNN models to knowledge graph-based data to enhance contextual information for event prediction by learning multi-relational features~\cite{deng2020dynamic,deng2021understanding}. 
For crime prediction, some work employed GNN models to enhance representation learning via capturing spatial correlations among pre-defined graph nodes, thereby improving forecasting accuracy~\cite{jin2020addressing,wang2020deep,wei2020crimestc,xiaspatial}.

\section{Open Challenges and Future Directions}\label{challenge}
Despite some achievements in societal event prediction in recent years, there are still open problems. In what follows, we discuss some challenges and future directions from three aspects.

\subsection{Data Dynamics, Sufficiency, and Reliability}
Data-driven approaches for societal event prediction depend heavily on data quality, which makes them subject to a number of data challenges.
The dynamic nature of data is one of the main challenges.
For text data, language, vocabulary, and mainstream slang are constantly evolving.
In geographic data, location names and area boundaries may change due to major political events.
Proposing advanced deep learning methods that can overcome the problem of data dynamics is of great significance for practical applications.
The sufficiency of data is another challenge. As seen in existing work, researchers have investigated various external data sources in addition to historical event occurrences to improve the accuracy of predictions.
Collecting external data from multiple sources and distinguishing correlated data from noisy data is expensive in terms of time, material, and computational costs.
Moreover, the spatial scarcity of societal events will also hinder event prediction studies in underrepresented areas.
A further fundamental issue in forecasting problems is data reliability.
Missing or incorrect data often occurs during manual or automated data collection.
For example, automated event collection systems may lose events due to unexpected network failures.
Social media posts provide valuable resources for tracking user behavior and social activities.
However, such data include typos, chit-chat, and misinformation that can mislead predictive models.
Hence, over-reliance on data can make prediction models vulnerable to real-world applications.

\subsection{Transparency and Interpretability of Deep Learning Models} 
Deep learning models have become increasingly sophisticated with increased predictive power. 
However, as model structures become more complex, the underlying mechanisms of predictive models become more opaque to humans, making model outputs more difficult to understand.
There is a growing awareness of the need for improving model transparency and interpretability.
Relying solely on model predictions to analyze societal events and make decisions can lead to economic loss or other unwanted consequences due to misjudgments.
One can mitigate such problems by providing reasonable explanations for model predictions so that practitioners can understand the behavior of the models and thus assess the reliability of the output produced by these models.
For societal event prediction, it is beneficial to develop explanatory models that can identify critical information from copious amounts of historical data.
Providing explanations for event prediction can improve the robustness of efficient event modeling supported by machine learning.

\subsection{Causality Study in Societal Events}
Machine learning methods generally focus on discovering correlations between input features and target variables rather than understanding causality.
Causality research or causal inference aims to study the cause and effect relationships (i.e., causality) between two or more variables. 
To better understand societal events and forecast these events with greater confidence, introducing causality into event modeling has potential benefits.
Analyzing probable causes of future events may allow us to develop causality-enhanced event prediction models that are not susceptible to data problems, such as noise and sparsity.
Investigating the causality of societal events can also help us reveal the underlying mechanisms or critical factors behind event occurrences in dynamic social environments.
A growing number of studies have absorbed the advantages of deep learning models, as well as causal studies~\cite{li2020teaching,chen2021spatial}.
The incorporation of causality in deep-learning based event predictions is becoming an indispensable research topic.

\section{Conclusion}\label{conclusion}

In this article, we present a comprehensive survey of existing methods for societal event forecasting, focusing on deep learning-based approaches.
We outline the challenges in societal event prediction.
We summarize research papers published in recent years and discuss data resources, research problems, traditional and deep learning predictive techniques for civil unrest and crime events, respectively.
In addition, we provide an overview of deep learning models that have been employed in prediction models, point out their characteristics and advantages in addressing different problems, and discuss open challenges and promising directions for future work.

% ...

  \bibliographystyle{ACM-Reference-Format}
  \bibliography{ref}
  
\end{document}